\documentclass{article} % For LaTeX2e
\usepackage{iclr2026_conference,times}

\usepackage{amsmath,amsfonts,bm}

\def\eqref#1{equation~\ref{#1}}
\def\1{\bm{1}}

\DeclareMathAlphabet{\mathsfit}{\encodingdefault}{\sfdefault}{m}{sl}
\SetMathAlphabet{\mathsfit}{bold}{\encodingdefault}{\sfdefault}{bx}{n}

\usepackage{hyperref}
\usepackage{url}
\usepackage{booktabs}
\usepackage{multirow}
\usepackage{graphicx}
\usepackage{makecell}
\usepackage[table]{xcolor}
\usepackage{algorithm}
\usepackage{algpseudocode}
\usepackage{subcaption}
\usepackage{enumitem}
\usepackage{siunitx}

\title{TableDART: Dynamic Adaptive Multi-Modal Routing for Table Understanding}

\author{\textbf{Xiaobo Xing}\textsuperscript{1*}, 
\textbf{Wei Yuan}\textsuperscript{1*}, 
\textbf{Tong Chen}\textsuperscript{1},
\textbf{Quoc Viet Hung Nguyen}\textsuperscript{2}, 
\textbf{Xiangliang Zhang}\textsuperscript{3},\\
\textbf{Hongzhi Yin}\textsuperscript{1$\dagger$}\\
\textsuperscript{1}The University of Queensland, Australia \\
\textsuperscript{2}Griffith University, Australia \quad \textsuperscript{3}University of Notre Dame, USA \\
\texttt{\{
\href{mailto:xiaobo.xing@uq.edu.au}{xiaobo.xing},
\href{mailto:w.yuan@uq.edu.au}{w.yuan},
\href{mailto:tong.chen@uq.edu.au}{tong.chen},
\href{mailto:h.yin1@uq.edu.au}{h.yin1}
\}@uq.edu.au} \\
\texttt{
\href{mailto:henry.nguyen@griffith.edu.au}{henry.nguyen@griffith.edu.au},
\href{mailto:xzhang33@nd.edu}{xzhang33@nd.edu}
}
}

\iclrfinalcopy % Uncomment for camera-ready version, but NOT for submission.
\begin{document}

\maketitle

% --- Start of Footnote Block ---
\begingroup
    \renewcommand\thefootnote{*}
    \footnotetext{Co-first authors with equal contributions. \quad 
    \textsuperscript{$\dagger$}Corresponding author.}
\endgroup
% --- End of Footnote Block ---

\begin{abstract}

Modeling semantic and structural information from tabular data remains a core challenge for effective table understanding. Existing Table-as-Text approaches flatten tables for large language models (LLMs), but lose crucial structural cues, while Table-as-Image methods preserve structure yet struggle with precise semantics. Recent Table-as-Multimodality strategies attempt to combine textual and visual views, but they (1) statically process both modalities for every query-table pair within large multimodal LLMs (MLLMs), inevitably introducing redundancy and even conflicts, and (2) depend on costly fine-tuning of MLLMs.  In light of this, we propose TableDART, a training-efficient framework that integrates multimodal views by reusing pretrained single-modality models.
TableDART introduces a lightweight 2.59M-parameter MLP gating network that dynamically selects the optimal path (Text-only, Image-only, or Fusion) for each table–query pair, reducing redundancy and avoiding conflicts that arise when textual and visual views of the same table provide inconsistent cues. By routing to the most appropriate view, our framework improves both accuracy and efficiency.
In addition, we propose a novel agent to mediate cross-modal knowledge integration by analyzing outputs from text- and image-based models, either selecting the best result or synthesizing a new answer through reasoning. This design avoids the prohibitive costs of full MLLM fine-tuning. Extensive experiments on seven benchmarks show that TableDART establishes new state-of-the-art performance among open-source models, surpassing the strongest baseline by an average of 4.02\%. The code is available at: \href{https://github.com/xiaobo-xing/TableDART}{https://github.com/xiaobo-xing/TableDART}.

\end{abstract}

\section{Introduction}

Tabular data is one of the most ubiquitous formats in the real world, with applications spanning finance \citep{DBLP:conf/emnlp/ChenCSSBLMBHRW21,DBLP:conf/acl/ZhuLHWZLFC20,DBLP:conf/icaif/ZhuLFWLC24}, healthcare \citep{DBLP:conf/emnlp/Shi0ZYZWZH0W24,DBLP:conf/ijcai/0010GX024}, and many other domains that rely on relational databases and spreadsheets \citep{DBLP:conf/icml/FeyHHLR0YYL24, gao2026relationaldatabasedistillationstructured}. Despite its prevalence, effectively modeling tabular data remains a long-standing challenge, due to its unique characteristics such as heterogeneity of diverse data types across columns, permutation invariance regarding row and column order, and hierarchical structure of multi-level or nested headers and indices
% \footnote{Respectively, these refer to the presence of diverse data types (e.g., numerical, categorical, and text) across columns; the property that the semantic meaning of the table is unchanged by shuffling its rows or columns; and the existence of multi-level or nested headers and indices.} 
\citep{DBLP:journals/tmlr/FangXTH0QSF24,DBLP:journals/tnn/BorisovLSHPK24,DBLP:conf/nips/ChenSLSZHK23}.

Existing approaches to tabular data understanding can be broadly categorized into two paradigms. The first is the ``\emph{Table-as-Text}" paradigm, where tables are linearized into sequences and processed by large language models (LLMs) \citep{zha2023tablegptunifyingtablesnature, su2024tablegpt2largemultimodalmodel}.
While effective, this approach is sensitive to serialization choices \citep{DBLP:conf/wsdm/SuiZZH024} and inevitably loses critical structural information \citep{DBLP:conf/acl/DengSHS0M0M24}.
The second paradigm treats ``\emph{Table-as-Image}", where table screenshots are processed with vision models. This approach better preserves structural information but struggles to capture precise and aligned semantic meaning \citep{DBLP:conf/acl/DengSHS0M0M24}.

To combine the complementary strengths of these paradigms, recent research has moved toward ``\emph{Table-as-Multimodality}'', where tables are represented in both textual and visual forms and processed via fine-tuned multimodal LLMs (MLLMs) \citep{liu2025hippoenhancingtableunderstanding}.
While promising, this MLLM-based paradigm suffers from two key limitations that hinder both performance and practicality.
First, regardless of the context, they mandate both modality views for every table-query pair. However, not all queries benefit from multiple views, as integrating multiple views often introduces redundancy and, in some cases, conflicting signals.
For example, textual linearization may inadvertently impose row-order sensitivity \citep{DBLP:journals/tmlr/FangXTH0QSF24}, while tables in their original or visual form remain permutation-invariant \citep{wu2025tabular}.
Such redundancy and conflicts mislead the MLLM's table understanding, resulting in suboptimal performance.
Second, existing approaches typically rely on heavy MLLMs, which remain computationally prohibitive to adapt at scale even with parameter-efficient fine-tuning strategies \citep{hu2022lora}, limiting their practicality under constrained training budgets and deployment pipelines \citep{DBLP:journals/dase/YinQCYZLXSZ25}.

In this paper, we propose TableDART (Dynamic Adaptive multi-modal RouTing), a novel framework for efficient and effective multimodal table understanding.
Unlike prior work, TableDART does not naively combine all modalities' perspectives. 
Instead, it introduces a lightweight MLP gating network that dynamically selects the optimal processing path (either Text-only, Image-only, or Fusion) for each table-query pair depending on the instance's complexity and resource efficiency.
When fusion is selected, a powerful LLM agent integrates the outputs of both single-modality experts, serving either as an arbitrator (selecting the better output) or as a rescuer (generating an enhanced answer by reasoning over both sources). Notably, TableDART reuses existing single-modality experts (Text-only and Image-only), requiring training only for the gating network. This makes it significantly more training-efficient than fine-tuning MLLMs.
We validate TableDART with extensive experiments on seven diverse benchmarks. Results show that TableDART not only achieves state-of-the-art performance among open-source models, surpassing the strongest MLLM-based baseline by an average of 4.02\% accuracy, but also provides new insights into dynamic routing policies for multimodal table understanding.

In summary, our main contributions are:
\begin{enumerate}[itemsep=0pt, topsep=0pt, label=(\alph*)]
\item  We propose TableDART, a novel framework that adaptively selects the optimal processing path for each query to make the most of modality-specific features for table understanding.
\item We introduce a parameter-efficient training paradigm that fully leverages pretrained (and frozen) single-modality experts by learning to route over experts with different output vocabularies, making the lightweight gating network (2.59M parameters) the only trainable part. An LLM agent is designed to further facilitate cross-modal consensus, enabling a plug-and-play setup without fine-tuning any LLM or MLLM backbones.
\item We conduct extensive experiments and establish new state-of-the-art performance on multiple benchmarks. In addition, we provide an in-depth analysis of the learned routing policy, demonstrating its strong generalizability and effectiveness.
\end{enumerate}

% Related Work Section
\section{Related Work}
% \textbf{Advancements in Large Language Models.}

\textbf{Unimodal Approaches for Table Understanding.}
A dominant paradigm for applying LLMs to tabular data is the Table-as-Text approach, which serializes tables into linear text sequences. Early work such as TaPas \citep{DBLP:conf/acl/HerzigNMPE20} demonstrated its effectiveness, inspiring specialized tabular models like TableLlama \citep{DBLP:conf/naacl/ZhangYL024} and TableGPT2 \citep{su2024tablegpt2largemultimodalmodel}, which achieve state-of-the-art results for table understanding. However, these models remain constrained by their input modality, as they cannot capture the rich visual and structural semantics inherent in the original table format \citep{DBLP:conf/acl/DengSHS0M0M24}. To overcome this limitation, the Table-as-Image paradigm has emerged, leveraging Multi-modal Large Language Models (MLLMs) to directly process table images. Pioneering works such as Table-LLaVA \citep{DBLP:conf/acl/ZhengFSS0J024}, TabPedia \citep{DBLP:conf/nips/ZhaoFLTWLWY0ZLH24}, SynTab-LLaVA \citep{DBLP:conf/cvpr/ZhouGWZWCX25}, and generalist MLLMs like Ovis2 \citep{lu2024ovisstructuralembeddingalignment} and Qwen2.5-VL \citep{DBLP:journals/corr/abs-2502-13923} showcase the potential of modeling visual layouts and structural cues, aligning with broader trends in vision-centric text understanding \citep{yuan2026integratingvisioncentrictextunderstanding, qiao2026textasvisionmeetssemanticids}. Yet, this paradigm is not a universal solution, as its effectiveness diminishes for tasks where reasoning relies heavily on text-centric pretraining knowledge \citep{DBLP:conf/acl/DengSHS0M0M24}.

\textbf{Multi-modal Approaches and Dynamic Routing.}
Building on the complementary strengths of unimodal representations, Table-as-Multimodality approaches have emerged. For example, HIPPO \citep{liu2025hippoenhancingtableunderstanding} fine-tunes an MLLM to jointly process both text and image representations of the table, achieving strong performance gains. However, these MLLM-based methods typically adopt a static, one-size-fits-all strategy, applying both modalities to every table–query pair. This rigid design can introduce redundancy or even conflict, particularly when a single modality would suffice. To address this gap, we propose TableDART, which introduces a dynamic routing mechanism that adaptively selects the optimal modality path for each instance. By learning an intelligent, instance-level policy, TableDART enables more efficient and robust multimodal reasoning, moving beyond the limitations of prior paradigms.

% Methodology Section
\section{Methodology}
\label{sec:methodology}

\subsection{Overview}
\begin{figure*}[t]
\centering
\includegraphics[width=\textwidth]{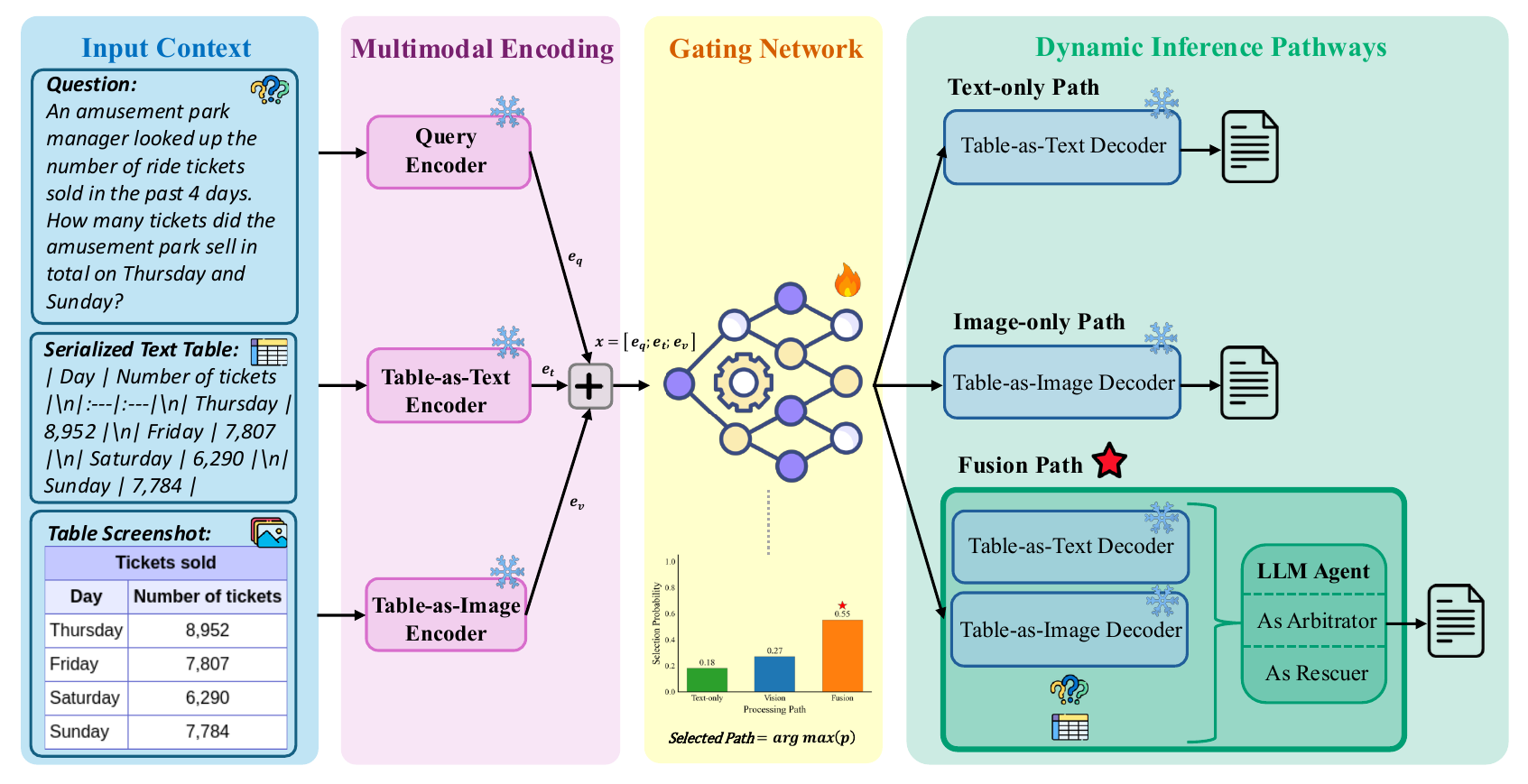}
\caption{
    \textbf{Architecture of TableDART.} The framework operates in three main stages: Multimodal Encoding (Section~\ref{sec:encoding_multimodality}), Gating Network (Section~\ref{sec:gating_training}), and Dynamic Inference Pathways (Section~\ref{sec:inference_multimodality}).
}
\label{fig:TableDART}
\vspace{-10pt}
\end{figure*}

The key novelty of TableDART lies in its ability to dynamically and efficiently leverage existing single-modality models to capture complementary information from tabular data for downstream tasks. To this end, TableDART integrates five components: (1) a Table-as-Text model $\mathcal{M}_{t}$, (2) a Table-as-Image model $\mathcal{M}_{v}$, (3) a text embedding model to process queries, (4) a lightweight gating network, and (5) an LLM-based agent responsible for multimodal fusion. Among these components, only the gating network is trainable, ensuring that TableDART remains highly parameter- and training-efficient.
Figure~\ref{fig:TableDART} provides an overview of TableDART's workflow, which is detailed in the subsequent sections.

\vspace{-10pt}

\subsection{Encoding Table-as-Multimodality}
\label{sec:encoding_multimodality}
Given a raw table $\mathcal{T}$ and a query $q$, TableDART leverages existing models to encode them from complementary modality perspectives, thereby capturing both semantic and structural information. 
This is achieved through a parallel feature extraction pipeline with three concurrent streams:
(1) Table-as-Text encoding: The raw table $\mathcal{T}$ is serialized into text and processed by the encoder $\mathcal{E}_{t}$ of the Table-as-Text model $\mathcal{M}_{t}$, yielding a feature vector $\mathbf{e}_{t}$.
(2) Table-as-Image encoding: A screenshot of $\mathcal{T}$ is fed into the encoder $\mathcal{E}_{v}$ of the Table-as-Image model $\mathcal{M}_{v}$, producing $\mathbf{e}_{v}$.
(3) Query encoding: The query $q$ is transformed into an embedding $\mathbf{e}_{q}$ using a text embedding model $\mathcal{E}_{q}$.
We apply modality-specific pooling to convert encoder outputs into fixed-dimensional vectors, which are concatenated to form a unified multimodal representation for the gating network:
\begin{equation}\label{eq_multi_encode}
\begin{aligned}
    \mathbf{e}_{t} = \mathcal{E}_{t}(\text{Serialize}(\mathcal{T}))\\
    \mathbf{e}_{v} = \mathcal{E}_{v}(\text{Screenshot}(\mathcal{T}))\\
    \mathbf{e}_{q} = \mathcal{E}_{q}(q)\\
    \mathbf{x} = [\mathbf{e}_{q}, \mathbf{e}_{t}, \mathbf{e}_{v}]
\end{aligned}
\end{equation}
Here, $\mathbf{x}$ serves as the multimodal joint representation by concatenating the three embeddings, which is subsequently passed into the gating network to determine the path of dynamic inference.
Note that $\mathcal{E}_{t}$ and $\mathcal{E}_{v}$ only activate a 
small part of the corresponding expert models, constituting a minor computational 
overhead (i.e., 7.15\% of the Table-as-Text model's and 7.63\% of the Table-as-Image 
model's total parameters). Detailed analyses of the embedding construction and 
parameter breakdown are provided in Appendix~\ref{appendix:embedding-analysis}.

\subsection{Gating Network and Policy Training}
\label{sec:gating_training}
As discussed in the introduction and confirmed by our experiments, not every table–query pair benefits from fusing both modalities, as this can introduce redundant or even conflicting signals. To address this, TableDART employs a lightweight gating network that dynamically selects the inference path. This design both maximizes the effective use of information and reduces computational cost compared to always invoking all modality models.

Formally, the gating network $\mathcal{G}$ is implemented by a lightweight MLP with few trainable parameters (see Appendix~\ref{appendix:gating-architecture} for architectural details). 
It takes as input the multimodal representation $\mathbf{x}$ from Eq.~\ref{eq_multi_encode} and outputs the raw logits $\mathbf{z}$ for each processing path (Text-only, Image-only, or Fusion):
\begin{equation}
\mathbf{z} = \mathcal{G}(\mathbf{x}).
\end{equation}

To ensure that $\mathcal{G}$ selects the most appropriate strategy for each table-query pair, balancing both predictive accuracy and resource efficiency, we design the following training objective:
\begin{equation}
\mathcal{L}_{\text{total}} = \mathcal{L}_{\text{task}} + \lambda \mathcal{L}_{\text{resource}},
\end{equation}
where $\lambda$ controls the strength of the resource-aware regularization.

The task loss $\mathcal{L}_{\text{task}}$ encourages the gating network to prioritize paths with strong empirical performance. For each training instance, we pre-compute a vector of binary scores $\mathbf{s} = (s_1, s_2, s_3) \in \{0, 1\}^3$, where each component $s_k$ is 1 if the $k$-th inference path produces a correct answer, and 0 otherwise. 
This method naturally handles instances where multiple paths are correct, as each successful path receives a score of 1. To form the training signal, this score vector is converted into a soft target probability distribution using a temperature-controlled softmax. Similarly, the gating network's raw logits $\mathbf{z}$ are converted into a predicted distribution. The task loss then minimizes the KL divergence between these two distributions:

\begin{equation}
\mathcal{L}_{\text{task}} = \text{KL}(\text{softmax}(\mathbf{s}/\tau) \parallel \text{softmax}(\mathbf{z}/\tau_g)),
\end{equation}
where $\tau_g, \tau$ are temperature parameters. We use the standard forward KL divergence, which encourages the gating network to assign probability to all empirically successful paths.

Beyond empirical performance on table understanding, we also account for inference overhead. To prevent over-reliance on costly strategies, we introduce a resource-aware regularizer:
\begin{equation}
\mathcal{L}_{\text{resource}} = \text{softmax}(\mathbf{z}/\tau_g)^T \mathbf{c},
\end{equation}
where $\mathbf{c}$ is an empirically measured cost vector for each inference path. By computing the expected cost of the routing policy, the resource loss $\mathcal{L}_{\text{resource}}$ penalizes high-cost inference routes, especially on simple tasks where $\mathcal{L}_{\text{task}}$ sees minimal improvements from over-complex paths. As such, it guides the gating policy toward more balanced choices, alleviating the information redundancy problem. The full training procedure, including the detailed protocol for measuring the cost vector $\mathbf{c}$, is provided in Appendix~\ref{appendix:experimental_setup_details}.

\subsection{Adaptive Inference}
\label{sec:inference_multimodality}
The adaptive inference process begins with a deterministic routing decision guided by the trained gating network. For each query-table pair, the network produces a vector of raw logit scores, and the framework selects the single, optimal path corresponding to the highest score for execution. 
Once the path is selected, inference proceeds accordingly. 
TableDART supports three inference paths: Text-only, Image-only, and Fusion. The first two directly reuse existing single-modality expert models: the Table-as-Text model $\mathcal{M}_{t}$ and the Table-as-Image model $\mathcal{M}_{v}$.
If the gating network selects either, TableDART simply continues the inference process from the intermediate representation of the chosen model. For example, when the gate selects the Text-only path, TableDART forwards $\mathcal{M}_{t}$'s remaining layers using the intermediate encoding $\mathbf{e}_{t}$ to produce the final output.

For the Fusion path, we design a novel, training-free LLM-based Fusion agent that synthesizes the final answer by reasoning over the outputs of the two single-modality experts, $\mathcal{M}_{t}$ and $\mathcal{M}_{v}$. Specifically, when the gating network selects the Fusion path, 
both $\mathcal{M}_{t}$ and $\mathcal{M}_{v}$ are at first executed to generate results $r_{t}$ and $r_{v}$, along with their auxiliary outputs $a_{t}$ and $a_{v}$. These, together with the original table $\mathcal{T}$, are passed to the Fusion agent.
The Fusion agent operates in two possible roles depending on its reasoning process:
(1) \textbf{Arbitrator}: If $\mathcal{M}_{t}$ and $\mathcal{M}_{v}$ produce conflicting results, the agent resolves the disagreement, similar in spirit to aggregating a consensus from multiple sources \citep{DBLP:journals/tkde/HungVTWYZ18}, by selecting the more reliable answer according to its confidence.
(2) \textbf{Rescuer}: If the agent believes both models provide uncertain or low-confidence outputs, it synthesizes a new, more accurate answer by jointly reasoning over their partial evidence.
A detailed description of the Fusion agent's implementation and prompt design is provided in Appendix~\ref{appendix:fusion-design}.

% Experiments Section
\section{Experiments}
\label{sec:experiments}

\subsection{Experimental Setup}

\textbf{Evaluation Datasets and Metrics.} We evaluate TableDART on seven diverse benchmarks across two tasks: Table Question Answering (TQA) and Table Fact Verification (TFV). For TQA, which requires interpreting complex queries \citep{DBLP:journals/tois/RenYCWHHZ20,DBLP:conf/sigir/RenYCWH021} to perform grounded reasoning over tabular evidence, we use five benchmarks: WTQ \citep{DBLP:conf/acl/PasupatL15}, TABMWP \citep{DBLP:conf/iclr/Lu0CWZRCK23}, TAT-QA \citep{DBLP:conf/acl/ZhuLHWZLFC20}, HiTab \citep{DBLP:conf/acl/Cheng0WJG0HLZ22}, and FeTaQA \citep{DBLP:journals/tacl/NanHMLVZKSKTMRT22}. For TFV, we use TabFact \citep{DBLP:conf/iclr/ChenWCZWLZW20} and InfoTabs \citep{DBLP:conf/acl/GuptaMNS20}.
Following the established protocol \citep{DBLP:conf/acl/ZhengFSS0J024,liu2025hippoenhancingtableunderstanding}, we use Accuracy for WTQ, TABMWP, TAT-QA, HiTab, TabFact, InfoTabs, while using BLEU score for FeTaQA as it is a generative free-form response task. Detailed statistics and further experimental setup details are provided in Appendix~\ref{appendix:experimental_setup_details}.

\textbf{Baselines.} We compare TableDART against a comprehensive set of baselines, including (1) Table-as-Text models: Llama-2-7B \citep{touvron2023llama}, Llama3-Instruct-8B \citep{dubey2024llama}, TableLlama-7B \citep{DBLP:conf/naacl/ZhangYL024}, and TableGPT2-7B \citep{su2024tablegpt2largemultimodalmodel}; (2) Table-as-Image models: MiniGPT-4-7B \citep{DBLP:conf/iclr/Zhu0SLE24}, mPLUG-Owl-7B \citep{ye2023mplug}, mPLUG-Owl2-7B \citep{ye2024mplug}, LLaVA v1.5-7B \citep{DBLP:conf/cvpr/LiuLLL24}, Table-LLaVA-7B \citep{DBLP:conf/acl/ZhengFSS0J024}, Qwen-VL-7B \citep{bai2023qwen}, InternLM-XComposer2-7B \citep{zhang2023internlm}, Monkey-7B \citep{DBLP:conf/cvpr/LiYLMZYSLB24}, TabPedia-7B \citep{DBLP:conf/nips/ZhaoFLTWLWY0ZLH24}, SynTab-LLaVA-7B \citep{DBLP:conf/cvpr/ZhouGWZWCX25}, MiniCPM-V-2.6-8B \citep{yao2024minicpm}, Qwen2.5-VL-7B \citep{DBLP:journals/corr/abs-2502-13923}, and Ovis2-8B \citep{lu2024ovisstructuralembeddingalignment}; (3) Table-as-Multimodality models: HIPPO-8B \citep{liu2025hippoenhancingtableunderstanding} and Google Gemini 2.0 Flash \citep{comanici2025gemini}. 

We employ TableGPT2-7B and Ovis2-8B as TableDART’s primary single-modality models. To demonstrate generalization, we additionally evaluate a variant by replacing the Table-as-Image model Ovis2-8B with Qwen2.5-VL-7B while retaining TableGPT2-7B. We use Google Gemini 2.0 Flash to implement the Fusion agent. Comparing TableDART against these base models directly demonstrates the effectiveness of our framework. 
Appendix~\ref{appendix:models_relationale} provides a detailed rationale for model selection and clarifies the sourcing of baseline results adopted from previous studies.
Notably, TableDART is a general framework that can be seamlessly integrated with a wide range of existing LLMs, VLMs and MLLMs. 

\textbf{Implementation Details.} \label{sec:implementation_details}
Our training set is a 10,000-sample mixture constructed by sampling from five diverse table understanding benchmarks, following the protocol of prior work \citep{liu2025hippoenhancingtableunderstanding}. We train only the lightweight MLP gating network while keeping all large LLM models frozen. The complete details regarding our data construction, hyperparameter settings, and computational environment are provided in Appendix~\ref{appendix:experimental_setup_details}.

\subsection{Effectiveness Analysis}
\label{sec:main_results}

\begin{table*}[t!]
\centering
\scriptsize
\caption{Main performance comparison on seven benchmarks for Table Question Answering (TQA) and Table Fact Verification (TFV). The Average column shows the mean accuracy across all benchmarks that use Accuracy as the evaluation metric. The \textbf{best} and \underline{second-best} results are marked.}
\label{tab:main-results}
\setlength{\tabcolsep}{3.9pt} 
\begin{tabular}{l ccccc cc c}
\toprule
% Row 1: Group Titles
& \multicolumn{5}{c}{\textbf{TQA}} & \multicolumn{2}{c}{\textbf{TFV}} & \multicolumn{1}{c}{\textbf{Summary}} \\
\cmidrule(lr){2-6} \cmidrule(lr){7-8} \cmidrule(lr){9-9}
% Row 2: Method and Dataset Names
\textbf{Method} & WTQ & TABMWP & TAT-QA & HiTab & FeTaQA & TabFact & InfoTabs & Average \\
% Row 3: Units
& (Acc.) & (Acc.) & (Acc.) & (Acc.) & (BLEU) & (Acc.) & (Acc.) & (Acc.) \\
\midrule
\rowcolor{gray!15} \multicolumn{9}{l}{\textit{Constituent Models of TableDART}} \\
TableGPT2-7B \textit{(Text-only Path)} & 61.42 & 83.87 & 50.39 & 70.27 & 28.97 & 77.80 & 71.07 & 69.14 \\
Ovis2-8B \textit{(Image-only Path)} & 58.76 & 87.00 & 47.67 & 68.59 & 34.70 & 80.80 & 74.11 & 69.49 \\
\midrule
\rowcolor{gray!15} \multicolumn{9}{l}{\textit{Table-as-Text Baselines}} \\
Llama-2-7B             & 16.39 & 22.82 & 13.73 & 10.72 & 10.93 & 9.20 & 38.92 & 18.63 \\
Llama3-Instruct-8B     & 21.24 & 42.01 & 13.08 & 6.97 & 12.66 & 73.89 & 54.00 & 35.20 \\
TableLlama-7B            & 24.97 & 10.10 & 19.04 & 46.57 & \textbf{38.38} & 79.37 & 46.57 & 37.77 \\
\rowcolor{gray!15} \multicolumn{9}{l}{\textit{Table-as-Image Baselines}} \\
MiniGPT-4-7B             & 0.90 & 0.22 & 0.13 & 0.20 & 0.39 & 0.00 & 0.10 & 0.26 \\
mPLUG-Owl-7B             & 0.62 & 1.76 & 0.13 & 0.25 & 7.42 & 7.46 & 5.53 & 2.63 \\
mPLUG-Owl2-7B            & 0.67 & 6.83 & 0.39 & 0.13 & 11.91 & 8.21 & 26.19 & 7.07 \\
LLaVA v1.5               & 1.24 & 6.05 & 2.97 & 2.03 & 8.24 & 18.90 & 28.31 & 9.92 \\
Table-LLaVA-7B           & 18.43 & 57.78 & 12.82 & 10.09 & 25.60 & 59.85 & 65.26 & 37.37 \\
Qwen-VL-7B               & 0.09 & 3.30 & 0.13 & 0.06 & 0.45 & 1.12 & 0.65 & 0.89 \\
InternLM-XComposer2-7B   & 0.05 & 0.06 & 0.26 & 0.12 & 2.62 & 1.19 & 1.11 & 0.46 \\
Monkey-7B                & 19.07 & 13.26 & 12.31 & 6.41 & 3.41 & 22.56 & 22.11 & 15.95 \\
TabPedia-7B              & 23.53 & 10.66 & 13.08 & 6.54 & 14.31 & 35.49 & 2.43 & 15.29 \\
SynTab-LLaVA-7B          & 39.59 & \textbf{88.30} & 51.94 & 35.66 & 35.45 & 70.78 & 69.42 & 59.28 \\
MiniCPM-V-2.6-8B         & 47.97 & 83.68 & 51.55 & 56.53 & 32.68 & 78.48 & 73.03 & 65.21 \\
Qwen2.5-VL-7B         & 54.37 & 63.69 & 51.94 & 62.69  & 10.99 & 75.81  & 70.13 & 63.11 \\
\rowcolor{gray!15} \multicolumn{9}{l}{\textit{Table-as-Multimodality (MLLM-based) Baselines}} \\
HIPPO-8B                 & 55.77 & \underline{87.50} & \underline{60.75} & 63.00 & 33.18 & \textbf{82.27} & \underline{75.74} & \underline{70.84} \\
Gemini 2.0 Flash         & 63.56 & 46.29 & 35.62 & 60.41 & 10.57 & 81.33 & 54.31 & 56.92 \\
\rowcolor{gray!15} \multicolumn{9}{l}{\textit{Table-as-Multimodality (Dynamic Adaptive Routing)}} \\
TableDART (TableGPT2-7B + Qwen2.5-VL-7B) & \underline{69.29} & 72.61 & 59.07 & \underline{71.13} & 29.87 & 77.94 & 71.46 & 70.25 \\
\textbf{TableDART (TableGPT2-7B + Ovis2-8B)}       & \textbf{70.58} & 84.54 & \textbf{62.05} & \textbf{74.37} & \underline{36.11} & 
\underline{81.37} & \textbf{76.22} & \textbf{74.86} \\
\bottomrule
\end{tabular}
\end{table*}
\vspace{-10pt}

\textbf{State-of-the-Art Performance.} Table~\ref{tab:main-results} shows the comparison results of TableDART with baselines.
From the results, we can observe that:
(1) Compared to all single-modality based methods, TableDART consistently achieves superior or highly competitive results, including surpassing its constituent models like TableGPT2-7B and Ovis2-8B, demonstrating that the framework successfully combines their strengths to achieve a result neither could reach alone.
(2) Compared to multi-modality based methods, TableDART's dynamic routing mechanism proves more effective than the MLLM-based HIPPO, outperforming it on five of the seven benchmarks.
(3) TableDART outperforms the powerful Fusion agent model (Google Gemini 2.0 Flash), confirming that the performance gains are driven by the intelligent routing mechanism, not just the capacity of the Fusion agent's backbone.
(4) TableDART demonstrates strong \textbf{generalization} across backbone models. When instantiated by replacing the Table-as-Image model with Qwen2.5-VL-7B while retaining TableGPT2-7B, the framework achieves a substantial +7.14\% average accuracy gain over the single-modality Qwen2.5-VL baseline, validating its model-agnostic effectiveness.
Furthermore, the `Average' column summarizes the average performance, showing that TableDART achieves the strongest results among all baselines, surpassing the best multi-modality tabular model, HIPPO-8B, by a decisive \textbf{+4.02\%} in average accuracy and \textbf{+2.93} points in BLEU score.

\textbf{Zero-Shot Generalization Performance.} To assess TableDART’s generalization capability, we compare it with the strongest MLLM-based baseline, HIPPO-8B, in a zero-shot setting on unseen datasets excluded from training. As shown in Table~\ref{tab:generalization}, TableDART demonstrates superior generalization. It achieves a \textbf{+18.05\%} accuracy improvement and a +2.93 BLEU score gain over HIPPO-8B. Notably, TableDART maintains nearly identical performance on seen (74.95\%) and unseen datasets (74.37\%), whereas HIPPO-8B suffers a sharp drop from 72.41\% to 63.00\%. This substantial advantage highlights that TableDART learns a genuinely generalizable routing policy, capable of adapting to new tables rather than overfitting to the training distribution.

\begin{table}[htbp]
\centering
\caption{Generalization performance on seen versus unseen datasets. Seen datasets were included in the training mixture, while unseen datasets were excluded from training.}
\label{tab:generalization}
\small
\begin{tabular}{lcc}
\toprule
\multirow{2}{*}{\textbf{Method}} & \textbf{Seen Datasets} & \textbf{Unseen Datasets} \\
& (Avg. Acc.) & (Avg. Acc. / BLEU) \\
\midrule
HIPPO-8B \textit{(Best MLLM-based Baseline)} & 72.41 & 63.00 / 33.18 \\
TableDART & 74.95 & 74.37 / 36.11 \\
\midrule
\textbf{Improvement} & \textbf{+3.51\%} & \textbf{+18.05\% / +2.93 pts} \\
\bottomrule
\end{tabular}
\vspace{-10pt}
\end{table}
\vspace{-10pt}

\subsection{Efficiency Analysis}

\textbf{Training Efficiency.} TableDART demonstrates exceptional training efficiency rooted in its architectural design. Our framework freezes all large constituent models, requiring only the training of a lightweight MLP gating network with just \textbf{2.59M} parameters. This contrasts sharply with MLLM-based approaches like HIPPO, which fine-tune their 8B backbone model using LoRA \citep{hu2022lora}. For their model architecture and LoRA configuration, the total number of trainable parameters reaches \textbf{25.87M}, meaning that HIPPO trains nearly \textbf{10 times more} parameters than our entire framework. This vast difference highlights the profound training efficiency and the plug-and-play scalability of our modular design.

\textbf{Inference Efficiency.} To quantify the practical benefits of dynamic adaptive routing, we benchmark TableDART's optimal configuration ($\lambda=0.15$) 
against a Non-Adaptive Fusion baseline that processes every instance via the full multimodal pipeline. Both settings use identical backbone models and Fusion agents, ensuring a fair comparison to isolate the efficiency gains of our adaptive policy.
The detailed evaluation protocol is in Appendix~\ref{appendix:efficiency_protocol}.
The results, presented in Table~\ref{tab:efficiency_breakdown}, demonstrate substantial efficiency gains. 
For example, TableDART achieves an average reduction of \textbf{24.5\%} in latency compared to the Non-Adaptive Fusion baseline, decreasing the mean inference time from 2.92s to 2.20s per sample. 
These improvements are a direct consequence of TableDART's learned routing policy, which reserves the computationally expensive Fusion path for instances that truly require multimodal reasoning, while assigning simpler cases to the more efficient unimodal paths.
As shown in Appendix~\ref{sec:appendix_detailed_analysis} and 
Appendix~\ref{sec:appendix_adaptive_behavior}, the model routes multimodally complex instances to the Fusion path while assigning simpler ones to the appropriate unimodal path. 
In particular, Figure~\ref{fig:complementarity_breakdown} shows that 34.8\% of TAT-QA instances cannot be solved by either single-modality model alone, and Figure~\ref{fig:routing_distribution} 
confirms that the learned policy routes 88.7\% of these instances to the powerful Fusion path.
In contrast, for simpler datasets such as TABMWP, where single-modality models are highly effective, the routing policy assigns 97.2\% of the instances to the more efficient Image-only path. 
Overall, TableDART’s inference efficiency derives from its intelligent dynamic routing mechanism, which preserves effectiveness while substantially reducing unnecessary computational cost.

\begin{table}[htbp]
\centering
\caption{Inference efficiency for TableDART with its Dynamic Adaptive Routing policy versus the Non-Adaptive Fusion baseline. TPS is short for Tokens per Second. ``$\downarrow$'' indicates lower is better, while ``$\uparrow$'' indicates higher is better.}
\label{tab:efficiency_breakdown}
\small
\setlength{\tabcolsep}{6pt} 
\begin{tabular}{l cc cc}
\toprule
& \multicolumn{2}{c}{\textbf{Dynamic Adaptive Routing}} & \multicolumn{2}{c}{\textbf{Non-Adaptive Fusion}} \\
\cmidrule(lr){2-3} \cmidrule(lr){4-5}
\textbf{Dataset} & Latency (s) $\downarrow$ & TPS $\uparrow$ & Latency (s) $\downarrow$ & TPS $\uparrow$ \\
\midrule
WTQ      & 2.39 & 8.90  & 2.84 & 1.33  \\
TABMWP   & 1.81 & 26.09 & 2.53 & 1.04  \\
TAT-QA   & 2.35 & 22.36 & 3.17 & 2.22  \\
HiTab    & 2.56 & 13.97 & 3.19 & 1.70  \\
FeTaQA   & 1.93 & 18.98 & 2.99 & 11.11 \\
TabFact  & 2.21 & 16.48 & 2.91 & 4.43  \\
InfoTabs & 2.17 & 17.60 & 2.78 & 5.09  \\
\midrule
\textbf{Average} & \textbf{2.20} & \textbf{17.77} & 2.92 & 3.85 \\
\bottomrule
\end{tabular}
\end{table}
\vspace{-10pt}

\subsection{Analysis of Each Inference Path}
\label{sec:analysis_architecture}
\begin{figure*}[t!]
    \centering
    \includegraphics[width=0.90\textwidth]{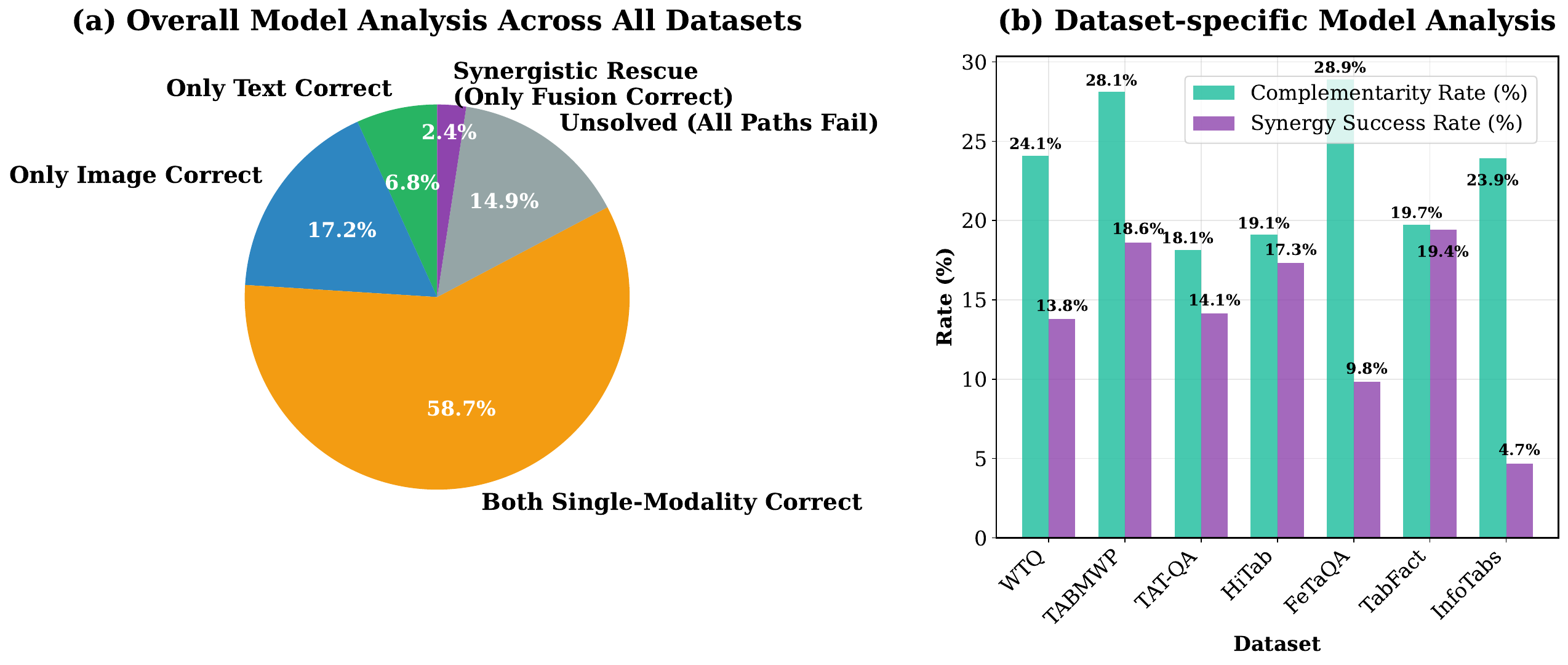} 
    \caption{Performance analysis of inference paths. (a) The chart counts instances based on which path(s), if any, produced a correct answer, across all datasets. (b) A per-dataset analysis of two key metrics: the Complementarity Rate, which is the percentage of instances where correctness is achieved by only one of the two single-modality models, and the Synergy Success Rate, which measures the fraction of hard cases (instances where both single-modality models fail) that are successfully resolved by the Fusion path.}
    \label{fig:expert_analysis_summary}
    \vspace{-10pt}
\end{figure*}

To better understand the contribution of each inference path in TableDART, we categorize instances based on whether the single-modality or Fusion paths could produce a correct solution. The statistics in Figure~\ref{fig:expert_analysis_summary} provide three key insights.
(1) Unimodal paths have complementary performance, as 24.0\% of test instances fall into a complementarity zone where only single-modality paths generate correct answers (17.2\% for the image-only path vs. 6.8\% for the text-only path) as shown in Figure~\ref{fig:expert_analysis_summary}(a). This underscores the importance of maintaining distinct single-modality inference paths.
Moreover, the complementarity rate varies across data sets according to Figure~\ref{fig:expert_analysis_summary}(b), highlighting the need for instance-specific dynamic routing.
(2) Hard cases, where both single-modality inference paths fail, account for 17.3\% of the test set. While 14.9\% of these remain unsolved, the Fusion path successfully rescues an additional 2.4\% (Figure~\ref{fig:expert_analysis_summary}a). This capability is further reflected in the Synergy Success Rate (i.e., the proportion of hard cases resolved by Fusion), which averages 14.0\% across benchmarks (Figure~\ref{fig:expert_analysis_summary}b). 
 These results confirm that our Fusion agent is not only an Arbitrator between modalities but also a Rescuer, capable of synergistic reasoning beyond the limits of individual models.
 (3) Easy cases where both unimodal inference paths succeed make up the majority of the data, accounting for 58.7\% of the test set (Figure~\ref{fig:expert_analysis_summary}a). For this substantial portion, multimodal fusion is unnecessary, as either single-modality path suffices to produce the correct answer. This result reinforces our core motivation for dynamic adaptive routing, rather than defaulting to multimodal fusion in all cases.

\vspace{-10pt}

\subsection{Ablation Study on Key Components}
\label{sec:ablation_mechanisms}

\textbf{Impact of Dynamic Gating.} 
We ablate the routing strategy to validate our dynamic approach by comparing it with two non-adaptive baselines in Table~\ref{tab:ablation_routing}. 
The results first confirm the effectiveness of our dynamic strategy, as TableDART significantly outperforms the Random Routing baseline. 
Compared with the Non-Adaptive Fusion baseline, which uses the same backbone models as TableDART but applies a brute-force strategy where every instance is routed through our LLM Agent-based Fusion path, our method not only achieves comparable results on most benchmarks but also delivers clear gains on datasets such as TABMWP and HiTab. 
This indicates that TableDART’s effectiveness is not solely attributable to the strong Fusion 
agent, but arises from the dynamic routing mechanism itself.
As shown in Figure~\ref{fig:routing_distribution}, the learned policy routes most instances to the 
appropriate unimodal paths and uses Fusion only when needed. In many cases, a single modality already provides sufficient information, and forcing unnecessary multimodal processing introduces extra cost and potential noise, which is why TableDART surpasses the Non-Adaptive Fusion baseline on several datasets.
Overall, these observations demonstrate the balance of resource-efficiency and performance-effectiveness of our dynamic routing design.

\begin{table}[htbp] \centering \small \setlength{\tabcolsep}{3.5pt} \begin{tabular}{l ccccc cc} \toprule & \multicolumn{5}{c}{\textbf{TQA}} & \multicolumn{2}{c}{\textbf{TFV}} \\ \cmidrule(lr){2-6} \cmidrule(lr){7-8} \textbf{Method} & WTQ & TABMWP & TAT-QA & HiTab & FeTaQA & TabFact & InfoTabs \\ \midrule TableDART w/ Random Routing & 65.40 & 75.50 & 58.94 & 70.49 & 30.87 & 79.50 & 69.57 \\ TableDART w/ Non-Adaptive Fusion & \textbf{70.97} & 81.47 & \textbf{63.34} & 73.35 & 34.82 & \textbf{81.56} & \textbf{76.83} \\ \midrule TableDART (Dynamic Routing) & 70.58 & \textbf{84.54} & 62.05 & \textbf{74.37} & \textbf{36.11} & 81.37 & 76.22 \\ \bottomrule \end{tabular} \caption{Ablation on routing strategies.} \label{tab:ablation_routing} \end{table}

\begin{table}[htbp]
\centering
\small
\begin{tabular}{l ccccc}
\toprule
\textbf{Benchmark} & $\lambda=1.0$ & $\lambda=0.15$ & $\lambda=0.1$ & $\lambda=0.05$ & $\lambda=0.00$ \\
\midrule
\rowcolor{gray!15} \multicolumn{6}{l}{\textit{Table Question Answering (TQA)}} \\
WTQ & 64.94 & 70.58 & 67.96 & \underline{71.80} & \textbf{71.96} \\
TABMWP & 83.86 & 84.54 & 84.79 & \underline{85.08} & \textbf{85.47} \\
TAT-QA & 61.79 & 62.05 & 63.21 & \underline{64.12} & \textbf{64.38} \\
HiTab & 73.79 & \textbf{74.37} & 72.14 & \underline{74.30} & 73.22 \\
FeTaQA & 35.28 & \textbf{36.11} & 35.37 & 34.34 & 35.87 \\
\midrule
\rowcolor{gray!15} \multicolumn{6}{l}{\textit{Table Fact Verification (TFV)}} \\
TabFact & 77.30 & 81.37 & 79.55 & \textbf{81.39} & 79.81 \\
InfoTabs & 70.13 & \textbf{76.22} & 70.44 & 73.59 & 74.19 \\
\midrule
\textbf{Avg. Acc.} & 71.97 & \underline{74.86} & 73.02 & \textbf{75.05} & 74.84 \\
\bottomrule
\end{tabular}
\caption{Impact of the resource loss weight ($\lambda$) on TableDART's performance. The \textbf{best} and \underline{second-best} results are highlighted.}
\label{tab:lambda_ablation}
\end{table}

\begin{figure*}[t!]
    \centering
    \includegraphics[width=0.90\textwidth]{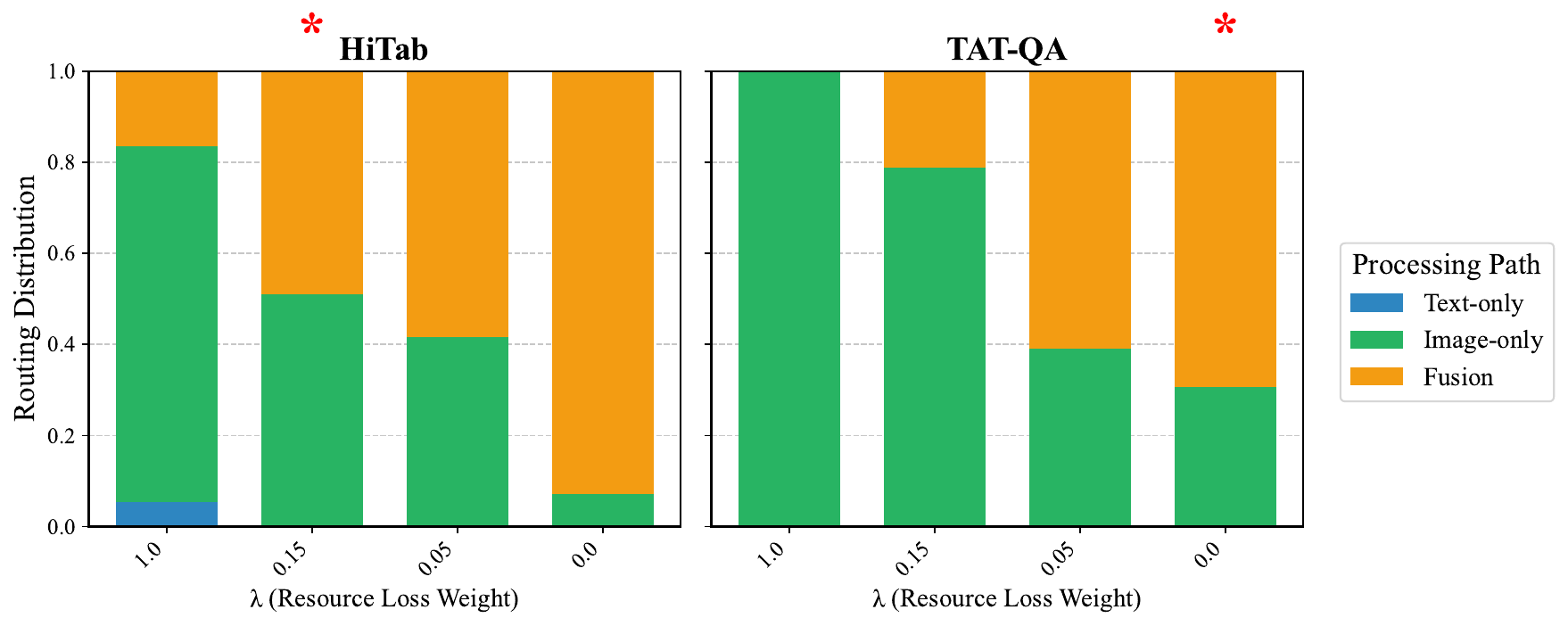}
    \caption{Inference path selection distribution vs. the resource loss weight ($\lambda$) on two representative benchmarks. Each bar shows the percentage of instances routed to the Text-only (blue), Image-only (green), and Fusion (orange) paths. A red star (*) marks the configuration with the highest performance for each dataset (see Table~\ref{tab:lambda_ablation}). This selection highlights TableDART's adaptability to diverse challenges. Full results on all seven datasets are provided in Appendix~\ref{sec:appendix_full_charts}.}
    \label{fig:main_body_chart}
    \vspace{-10pt}
\end{figure*}

\textbf{Impact of Resource-Aware Objective ($\lambda$).}
We analyze the effect of the resource loss weight $\lambda$, highlighting its two key roles in shaping the final routing policy. (1) \textbf{Effective Routing Policy Control.} As shown in Figure~\ref{fig:main_body_chart}, $\lambda$ directly regulates the framework’s behavior. For instance, reducing its value increases reliance on the computationally expensive Fusion path, causing more queries to be routed to the Fusion inference path.

(2) \textbf{Regularization for Improved Generalization.}   Table~\ref{tab:lambda_ablation} shows that our proposed resource-aware objective also serves as an effective regularizer. The highest performance, both in terms of average results and across four of the seven benchmarks, is not attained with a purely performance-driven policy ($\lambda=0.00$), but rather at a non-zero $\lambda$. By penalizing computational cost, the objective discourages the gate from over-relying on the Fusion path, a tendency that risks overfitting to training artifacts. Instead, it promotes a more balanced routing policy that generalizes more effectively. 
In addition, varying $\lambda$ has a direct impact on inference efficiency. As illustrated in Appendix~\ref{sec:appendix_lambda_efficiency} (Figure~\ref{fig:performance_efficiency_tradeoff} and Table~\ref{tab:lambda_efficiency}), efficiency generally improves as $\lambda$ 
increases, since a larger penalty discourages unnecessary use of the Fusion path. However, the trend is not strictly monotonic, because each $\lambda$ induces a distinct routing policy that 
balances accuracy and computational cost differently.
Based on this observation, we select $\lambda=0.15$ for the final model, which achieves the second-best 
average accuracy (74.86\%, within 0.19 points of the best) while requiring 8.4\% less inference latency, representing a well-regularized operating point on the performance–efficiency frontier.
This configuration enables a truly adaptive policy, which tailors routing to the task type, directing 97.2\% of queries to the Image-only path for the structure-heavy TABMWP, while shifting to 67.5\% Text-only for the semantics-focused InfoTabs (see Appendix~\ref{sec:appendix_adaptive_behavior} and Fig.~\ref{fig:routing_distribution} for a full breakdown).

\vspace{-10pt}

\subsection{Case Study}
\label{sec:case_study}

\begin{figure}[t]
    \centering
    \includegraphics[width=\columnwidth]{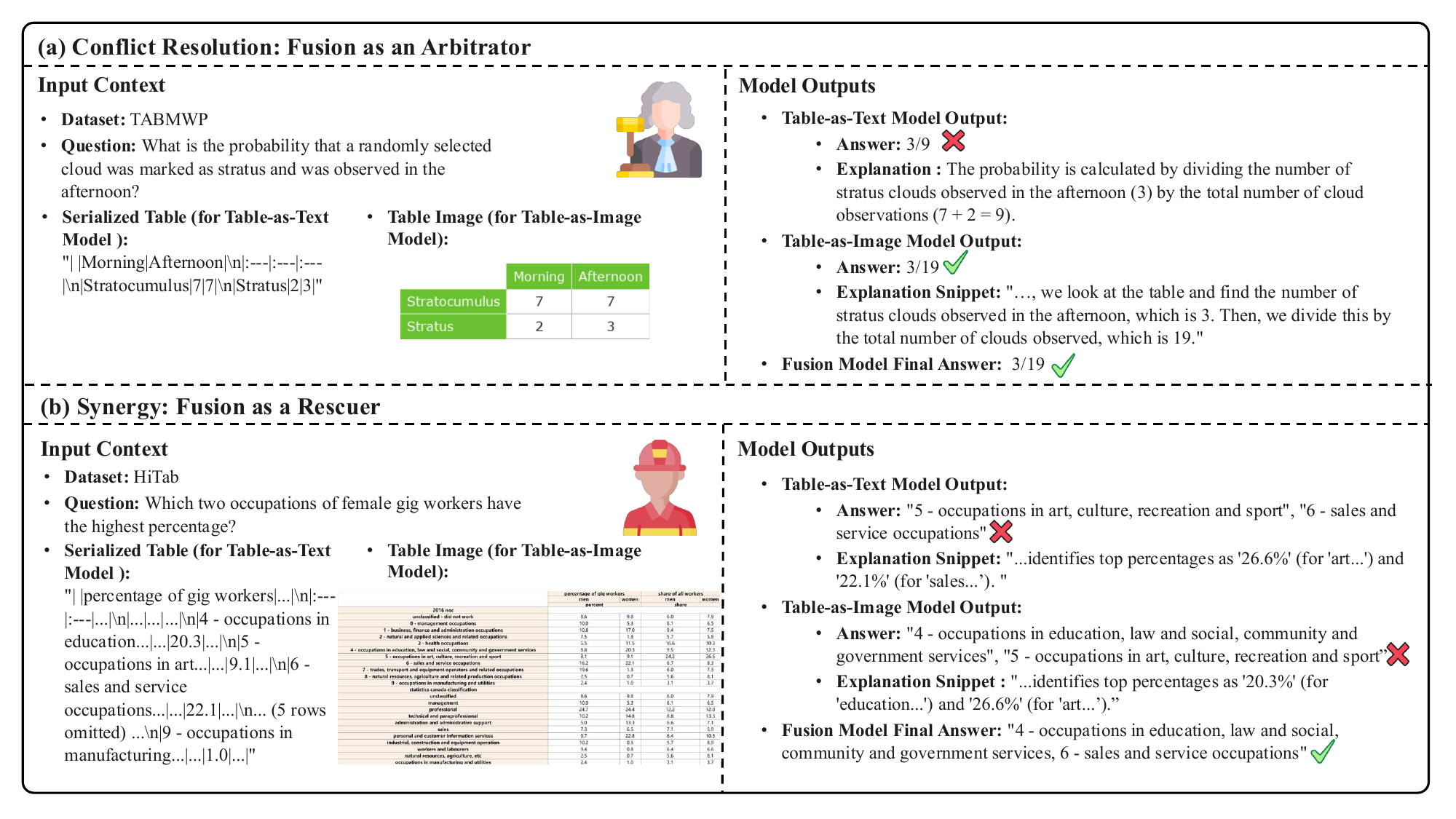}
    \caption{Case studies illustrate the key synthesis roles of the Fusion Model, which is implemented by an LLM agent. (a) As an \textbf{Arbitrator} (example from the TABMWP dataset), it resolves a conflict between a correct and an incorrect numerical reasoning path. (b) As a \textbf{Rescuer} (example from the HiTab dataset), it demonstrates synergy by synthesizing a correct answer from two distinct, incorrect outputs, showcasing its ability to combine partially correct reasoning fragments.}
    \label{fig:fusion-cases}
    \vspace{-10pt}
\end{figure}

To qualitatively illustrate the sophisticated reasoning of our Fusion path, Figure~\ref{fig:fusion-cases} presents two representative cases that highlight the LLM agent's primary roles. (1) The first case demonstrates the agent as an Arbitrator. Tasked with a probability calculation, the Table-as-Text model fails by incorrectly summing a subset of the data for its denominator, while the Table-as-Image model reasons correctly. The Fusion agent resolves this conflict by validating both reasoning paths against the source table and selecting the correct output. (2) The second case showcases the Fusion agent as a Rescuer. Here, Fusion agent reasons that both single-modality models fail to identify the two required occupations, with each providing only one correct and one hallucinated answer. The agent exhibits true synergy by synthesizing a new, fully correct answer, combining the valid fragments from both failed outputs while discarding the errors. This synthesis demonstrates the Fusion agent's ability to generate novel, correct answers from incomplete and conflicting information from two single-modality models.

\vspace{-10pt}

\section{Conclusion}
\label{sec:conclusion}

In this work, we introduced TableDART, a dynamic adaptive multi-modal routing framework for table understanding. We replace the conventional MLLM-based approach with a lightweight MLP-based gating network that learns to route each input to an optimal processing path: Text-only, Image-only, or Fusion. This instance-level routing is learned efficiently by training only the 2.59M-parameter gate while keeping the large base models entirely frozen, preserving their pre-trained capabilities and ensuring plug-and-play flexibility. Our experiments across seven diverse benchmarks demonstrate that TableDART consistently achieves strong performance, validating that its dynamic, resource-aware routing is a more effective and efficient paradigm for complex table reasoning tasks. Please refer to Appendix~\ref{appendix:reproducibility} for a detailed reproducibility statement.

\section*{Acknowledgments}
The Australian Research Council partially supports this work under the streams of Future Fellowship (Grant No. FT210100624), the Discovery Project (Grant No. DP240101108 and DP260100326), and the Linkage Project (Grant No. LP230200892 and LP240200546).

\bibliography{iclr2026_conference}

@article{hu2022lora,
  title={Lora: Low-rank adaptation of large language models.},
  author={Hu, Edward J and Shen, Yelong and Wallis, Phillip and Allen-Zhu, Zeyuan and Li, Yuanzhi and Wang, Shean and Wang, Lu and Chen, Weizhu and others},
  journal={ICLR},
  volume={1},
  number={2},
  pages={3},
  year={2022}
}

@inproceedings{DBLP:conf/acl/DengSHS0M0M24,
  author       = {Naihao Deng and
                  Zhenjie Sun and
                  Ruiqi He and
                  Aman Sikka and
                  Yulong Chen and
                  Lin Ma and
                  Yue Zhang and
                  Rada Mihalcea},
  title        = {Tables as Texts or Images: Evaluating the Table Reasoning Ability
                  of LLMs and MLLMs},
  booktitle    = {{ACL} (Findings)},
  pages        = {407--426},
  publisher    = {Association for Computational Linguistics},
  year         = {2024}
}

@inproceedings{DBLP:conf/acl/ZhengFSS0J024,
  author       = {Mingyu Zheng and
                  Xinwei Feng and
                  Qingyi Si and
                  Qiaoqiao She and
                  Zheng Lin and
                  Wenbin Jiang and
                  Weiping Wang},
  title        = {Multimodal Table Understanding},
  booktitle    = {{ACL} {(1)}},
  pages        = {9102--9124},
  publisher    = {Association for Computational Linguistics},
  year         = {2024}
}

@misc{jiang2025multimodaltabularreasoningprivileged,
      title={Multimodal Tabular Reasoning with Privileged Structured Information}, 
      author={Jun-Peng Jiang and Yu Xia and Hai-Long Sun and Shiyin Lu and Qing-Guo Chen and Weihua Luo and Kaifu Zhang and De-Chuan Zhan and Han-Jia Ye},
      year={2025},
      eprint={2506.04088},
      archivePrefix={arXiv},
      primaryClass={cs.LG},
      url={https://arxiv.org/abs/2506.04088}, 
}

@misc{liu2025hippoenhancingtableunderstanding,
      title={HIPPO: Enhancing the Table Understanding Capability of Large Language Models through Hybrid-Modal Preference Optimization}, 
      author={Zhenghao Liu and Haolan Wang and Xinze Li and Qiushi Xiong and Xiaocui Yang and Yu Gu and Yukun Yan and Qi Shi and Fangfang Li and Ge Yu and Maosong Sun},
      year={2025},
      eprint={2502.17315},
      archivePrefix={arXiv},
      primaryClass={cs.CL},
      url={https://arxiv.org/abs/2502.17315}, 
}

@inproceedings{DBLP:conf/icaif/ZhuLFWLC24,
  author       = {Fengbin Zhu and
                  Ziyang Liu and
                  Fuli Feng and
                  Chao Wang and
                  Moxin Li and
                  Tat{-}Seng Chua},
  title        = {{TAT-LLM:} {A} Specialized Language Model for Discrete Reasoning over
                  Financial Tabular and Textual Data},
  booktitle    = {{ICAIF}},
  pages        = {310--318},
  publisher    = {{ACM}},
  year         = {2024}
}

@inproceedings{DBLP:conf/emnlp/Shi0ZYZWZH0W24,
  author       = {Wenqi Shi and
                  Ran Xu and
                  Yuchen Zhuang and
                  Yue Yu and
                  Jieyu Zhang and
                  Hang Wu and
                  Yuanda Zhu and
                  Joyce C. Ho and
                  Carl Yang and
                  May Dongmei Wang},
  title        = {EHRAgent: Code Empowers Large Language Models for Few-shot Complex
                  Tabular Reasoning on Electronic Health Records},
  booktitle    = {{EMNLP}},
  pages        = {22315--22339},
  publisher    = {Association for Computational Linguistics},
  year         = {2024}
}

@inproceedings{DBLP:conf/emnlp/ChenCSSBLMBHRW21,
  author       = {Zhiyu Chen and
                  Wenhu Chen and
                  Charese Smiley and
                  Sameena Shah and
                  Iana Borova and
                  Dylan Langdon and
                  Reema Moussa and
                  Matt Beane and
                  Ting{-}Hao Kenneth Huang and
                  Bryan R. Routledge and
                  William Yang Wang},
  title        = {FinQA: {A} Dataset of Numerical Reasoning over Financial Data},
  booktitle    = {{EMNLP} {(1)}},
  pages        = {3697--3711},
  publisher    = {Association for Computational Linguistics},
  year         = {2021}
}

@inproceedings{DBLP:conf/ijcai/0010GX024,
  author       = {Zifeng Wang and
                  Chufan Gao and
                  Cao Xiao and
                  Jimeng Sun},
  title        = {MediTab: Scaling Medical Tabular Data Predictors via Data Consolidation,
                  Enrichment, and Refinement},
  booktitle    = {{IJCAI}},
  pages        = {6062--6070},
  publisher    = {ijcai.org},
  year         = {2024}
}

@inproceedings{DBLP:conf/icml/FeyHHLR0YYL24,
  author       = {Matthias Fey and
                  Weihua Hu and
                  Kexin Huang and
                  Jan Eric Lenssen and
                  Rishabh Ranjan and
                  Joshua Robinson and
                  Rex Ying and
                  Jiaxuan You and
                  Jure Leskovec},
  title        = {Position: Relational Deep Learning - Graph Representation Learning
                  on Relational Databases},
  booktitle    = {{ICML}},
  publisher    = {OpenReview.net},
  year         = {2024}
}

@article{DBLP:journals/tmlr/FangXTH0QSF24,
  author       = {Xi Fang and
                  Weijie Xu and
                  Fiona Anting Tan and
                  Ziqing Hu and
                  Jiani Zhang and
                  Yanjun Qi and
                  Srinivasan H. Sengamedu and
                  Christos Faloutsos},
  title        = {Large Language Models (LLMs) on Tabular Data: Prediction, Generation,
                  and Understanding - {A} Survey},
  journal      = {Trans. Mach. Learn. Res.},
  volume       = {2024},
  year         = {2024}
}

@article{DBLP:journals/tnn/BorisovLSHPK24,
  author       = {Vadim Borisov and
                  Tobias Leemann and
                  Kathrin Se{\ss}ler and
                  Johannes Haug and
                  Martin Pawelczyk and
                  Gjergji Kasneci},
  title        = {Deep Neural Networks and Tabular Data: {A} Survey},
  journal      = {{IEEE} Trans. Neural Networks Learn. Syst.},
  volume       = {35},
  number       = {6},
  pages        = {7499--7519},
  year         = {2024}
}

@inproceedings{DBLP:conf/nips/ChenSLSZHK23,
  author       = {Pei Chen and
                  Soumajyoti Sarkar and
                  Leonard Lausen and
                  Balasubramaniam Srinivasan and
                  Sheng Zha and
                  Ruihong Huang and
                  George Karypis},
  title        = {HyTrel: Hypergraph-enhanced Tabular Data Representation Learning},
  booktitle    = {NeurIPS},
  year         = {2023}
}

@misc{zha2023tablegptunifyingtablesnature,
      title={TableGPT: Towards Unifying Tables, Nature Language and Commands into One GPT}, 
      author={Liangyu Zha and Junlin Zhou and Liyao Li and Rui Wang and Qingyi Huang and Saisai Yang and Jing Yuan and Changbao Su and Xiang Li and Aofeng Su and Tao Zhang and Chen Zhou and Kaizhe Shou and Miao Wang and Wufang Zhu and Guoshan Lu and Chao Ye and Yali Ye and Wentao Ye and Yiming Zhang and Xinglong Deng and Jie Xu and Haobo Wang and Gang Chen and Junbo Zhao},
      year={2023},
      eprint={2307.08674},
      archivePrefix={arXiv},
      primaryClass={cs.AI},
      url={https://arxiv.org/abs/2307.08674}, 
}

@misc{su2024tablegpt2largemultimodalmodel,
      title={TableGPT2: A Large Multimodal Model with Tabular Data Integration}, 
      author={Aofeng Su and Aowen Wang and Chao Ye and Chen Zhou and Ga Zhang and Gang Chen and Guangcheng Zhu and Haobo Wang and Haokai Xu and Hao Chen and Haoze Li and Haoxuan Lan and Jiaming Tian and Jing Yuan and Junbo Zhao and Junlin Zhou and Kaizhe Shou and Liangyu Zha and Lin Long and Liyao Li and Pengzuo Wu and Qi Zhang and Qingyi Huang and Saisai Yang and Tao Zhang and Wentao Ye and Wufang Zhu and Xiaomeng Hu and Xijun Gu and Xinjie Sun and Xiang Li and Yuhang Yang and Zhiqing Xiao},
      year={2024},
      eprint={2411.02059},
      archivePrefix={arXiv},
      primaryClass={cs.LG},
      url={https://arxiv.org/abs/2411.02059}, 
}

@inproceedings{DBLP:conf/wsdm/SuiZZH024,
  author       = {Yuan Sui and
                  Mengyu Zhou and
                  Mingjie Zhou and
                  Shi Han and
                  Dongmei Zhang},
  title        = {Table Meets {LLM:} Can Large Language Models Understand Structured
                  Table Data? {A} Benchmark and Empirical Study},
  booktitle    = {{WSDM}},
  pages        = {645--654},
  publisher    = {{ACM}},
  year         = {2024}
}

@inproceedings{DBLP:conf/iclr/LepikhinLXCFHKS21,
  author       = {Dmitry Lepikhin and
                  HyoukJoong Lee and
                  Yuanzhong Xu and
                  Dehao Chen and
                  Orhan Firat and
                  Yanping Huang and
                  Maxim Krikun and
                  Noam Shazeer and
                  Zhifeng Chen},
  title        = {GShard: Scaling Giant Models with Conditional Computation and Automatic
                  Sharding},
  booktitle    = {{ICLR}},
  publisher    = {OpenReview.net},
  year         = {2021}
}

@inproceedings{DBLP:conf/cvpr/ZhuZLWLWD22,
  author       = {Xizhou Zhu and
                  Jinguo Zhu and
                  Hao Li and
                  Xiaoshi Wu and
                  Hongsheng Li and
                  Xiaohua Wang and
                  Jifeng Dai},
  title        = {Uni-Perceiver: Pre-training Unified Architecture for Generic Perception
                  for Zero-shot and Few-shot Tasks},
  booktitle    = {{CVPR}},
  pages        = {16783--16794},
  publisher    = {{IEEE}},
  year         = {2022}
}

@inproceedings{DBLP:conf/cvpr/XueM23,
  author       = {Zihui Xue and
                  Radu Marculescu},
  title        = {Dynamic Multimodal Fusion},
  booktitle    = {{CVPR} Workshops},
  pages        = {2575--2584},
  publisher    = {{IEEE}},
  year         = {2023}
}

@misc{lu2024ovisstructuralembeddingalignment,
      title={Ovis: Structural Embedding Alignment for Multimodal Large Language Model}, 
      author={Shiyin Lu and Yang Li and Qing-Guo Chen and Zhao Xu and Weihua Luo and Kaifu Zhang and Han-Jia Ye},
      year={2024},
      eprint={2405.20797},
      archivePrefix={arXiv},
      primaryClass={cs.CV},
      url={https://arxiv.org/abs/2405.20797}, 
}

@article{comanici2025gemini,
  title={Gemini 2.5: Pushing the frontier with advanced reasoning, multimodality, long context, and next generation agentic capabilities},
  author={Comanici, Gheorghe and Bieber, Eric and Schaekermann, Mike and Pasupat, Ice and Sachdeva, Noveen and Dhillon, Inderjit and Blistein, Marcel and Ram, Ori and Zhang, Dan and Rosen, Evan and others},
  journal={arXiv preprint arXiv:2507.06261},
  year={2025}
}

@inproceedings{DBLP:conf/emnlp/ReimersG19,
  author       = {Nils Reimers and
                  Iryna Gurevych},
  title        = {Sentence-BERT: Sentence Embeddings using Siamese BERT-Networks},
  booktitle    = {{EMNLP/IJCNLP} {(1)}},
  pages        = {3980--3990},
  publisher    = {Association for Computational Linguistics},
  year         = {2019}
}

@article{dubey2024llama,
  title={The llama 3 herd of models},
  author={Dubey, Abhimanyu and Jauhri, Abhinav and Pandey, Abhinav and Kadian, Abhishek and Al-Dahle, Ahmad and Letman, Aiesha and Mathur, Akhil and Schelten, Alan and Yang, Amy and Fan, Angela and others},
  journal={arXiv e-prints},
  pages={arXiv--2407},
  year={2024}
}

@inproceedings{DBLP:conf/naacl/ZhangYL024,
  author       = {Tianshu Zhang and
                  Xiang Yue and
                  Yifei Li and
                  Huan Sun},
  title        = {TableLlama: Towards Open Large Generalist Models for Tables},
  booktitle    = {{NAACL-HLT}},
  pages        = {6024--6044},
  publisher    = {Association for Computational Linguistics},
  year         = {2024}
}

@inproceedings{DBLP:conf/iclr/Zhu0SLE24,
  author       = {Deyao Zhu and
                  Jun Chen and
                  Xiaoqian Shen and
                  Xiang Li and
                  Mohamed Elhoseiny},
  title        = {MiniGPT-4: Enhancing Vision-Language Understanding with Advanced Large
                  Language Models},
  booktitle    = {{ICLR}},
  publisher    = {OpenReview.net},
  year         = {2024}
}

@article{ye2023mplug,
  title={mplug-owl: Modularization empowers large language models with multimodality},
  author={Ye, Qinghao and Xu, Haiyang and Xu, Guohai and Ye, Jiabo and Yan, Ming and Zhou, Yiyang and Wang, Junyang and Hu, Anwen and Shi, Pengcheng and Shi, Yaya and others},
  journal={arXiv preprint arXiv:2304.14178},
  year={2023}
}

@inproceedings{ye2024mplug,
  title={mplug-owl2: Revolutionizing multi-modal large language model with modality collaboration},
  author={Ye, Qinghao and Xu, Haiyang and Ye, Jiabo and Yan, Ming and Hu, Anwen and Liu, Haowei and Qian, Qi and Zhang, Ji and Huang, Fei},
  booktitle={Proceedings of the ieee/cvf conference on computer vision and pattern recognition},
  pages={13040--13051},
  year={2024}
}

@inproceedings{DBLP:conf/cvpr/LiuLLL24,
  author       = {Haotian Liu and
                  Chunyuan Li and
                  Yuheng Li and
                  Yong Jae Lee},
  title        = {Improved Baselines with Visual Instruction Tuning},
  booktitle    = {{CVPR}},
  pages        = {26286--26296},
  publisher    = {{IEEE}},
  year         = {2024}
}

@article{bai2023qwen,
  title={Qwen technical report},
  author={Bai, Jinze and Bai, Shuai and Chu, Yunfei and Cui, Zeyu and Dang, Kai and Deng, Xiaodong and Fan, Yang and Ge, Wenbin and Han, Yu and Huang, Fei and others},
  journal={arXiv preprint arXiv:2309.16609},
  year={2023}
}

@article{zhang2023internlm,
  title={Internlm-xcomposer: A vision-language large model for advanced text-image comprehension and composition},
  author={Zhang, Pan and Dong, Xiaoyi and Wang, Bin and Cao, Yuhang and Xu, Chao and Ouyang, Linke and Zhao, Zhiyuan and Duan, Haodong and Zhang, Songyang and Ding, Shuangrui and others},
  journal={arXiv preprint arXiv:2309.15112},
  year={2023}
}

@inproceedings{DBLP:conf/cvpr/LiYLMZYSLB24,
  author       = {Zhang Li and
                  Biao Yang and
                  Qiang Liu and
                  Zhiyin Ma and
                  Shuo Zhang and
                  Jingxu Yang and
                  Yabo Sun and
                  Yuliang Liu and
                  Xiang Bai},
  title        = {Monkey: Image Resolution and Text Label are Important Things for Large
                  Multi-Modal Models},
  booktitle    = {{CVPR}},
  pages        = {26753--26763},
  publisher    = {{IEEE}},
  year         = {2024}
}

@article{yao2024minicpm,
  title={MiniCPM-V: A GPT-4V Level MLLM on Your Phone},
  author={Yao, Yuan and Yu, Tianyu and Zhang, Ao and Wang, Chongyi and Cui, Junbo and Zhu, Hongji and Cai, Tianchi and Li, Haoyu and Zhao, Weilin and He, Zhihui and others},
  journal={arXiv preprint arXiv:2408.01800},
  year={2024}
}

@article{DBLP:journals/csur/GuidottiMRTGP19,
  author       = {Riccardo Guidotti and
                  Anna Monreale and
                  Salvatore Ruggieri and
                  Franco Turini and
                  Fosca Giannotti and
                  Dino Pedreschi},
  title        = {A Survey of Methods for Explaining Black Box Models},
  journal      = {{ACM} Comput. Surv.},
  volume       = {51},
  number       = {5},
  pages        = {93:1--93:42},
  year         = {2019}
}

@inproceedings{DBLP:conf/acl/HerzigNMPE20,
  author       = {Jonathan Herzig and
                  Pawel Krzysztof Nowak and
                  Thomas M{\"{u}}ller and
                  Francesco Piccinno and
                  Julian Martin Eisenschlos},
  title        = {TaPas: Weakly Supervised Table Parsing via Pre-training},
  booktitle    = {{ACL}},
  pages        = {4320--4333},
  publisher    = {Association for Computational Linguistics},
  year         = {2020}
}

@article{touvron2023llama,
  title={Llama 2: Open foundation and fine-tuned chat models},
  author={Touvron, Hugo and Martin, Louis and Stone, Kevin and Albert, Peter and Almahairi, Amjad and Babaei, Yasmine and Bashlykov, Nikolay and Batra, Soumya and Bhargava, Prajjwal and Bhosale, Shruti and others},
  journal={arXiv preprint arXiv:2307.09288},
  year={2023}
}

@article{wu2025tabular,
  title={Tabular Data Understanding with LLMs: A Survey of Recent Advances and Challenges},
  author={Wu, Xiaofeng and Ritter, Alan and Xu, Wei},
  journal={arXiv preprint arXiv:2508.00217},
  year={2025}
}

@inproceedings{DBLP:journals/jmlr/GlorotB10,
  author       = {Xavier Glorot and
                  Yoshua Bengio},
  title        = {Understanding the difficulty of training deep feedforward neural networks},
  booktitle    = {{AISTATS}},
  series       = {{JMLR} Proceedings},
  volume       = {9},
  pages        = {249--256},
  publisher    = {JMLR.org},
  year         = {2010}
}

@article{DBLP:journals/tacl/NanHMLVZKSKTMRT22,
  author       = {Linyong Nan and
                  Chiachun Hsieh and
                  Ziming Mao and
                  Xi Victoria Lin and
                  Neha Verma and
                  Rui Zhang and
                  Wojciech Kryscinski and
                  Hailey Schoelkopf and
                  Riley Kong and
                  Xiangru Tang and
                  Mutethia Mutuma and
                  Ben Rosand and
                  Isabel Trindade and
                  Renusree Bandaru and
                  Jacob Cunningham and
                  Caiming Xiong and
                  Dragomir R. Radev},
  title        = {FeTaQA: Free-form Table Question Answering},
  journal      = {Trans. Assoc. Comput. Linguistics},
  volume       = {10},
  pages        = {35--49},
  year         = {2022}
}

@inproceedings{DBLP:conf/iclr/Lu0CWZRCK23,
  author       = {Pan Lu and
                  Liang Qiu and
                  Kai{-}Wei Chang and
                  Ying Nian Wu and
                  Song{-}Chun Zhu and
                  Tanmay Rajpurohit and
                  Peter Clark and
                  Ashwin Kalyan},
  title        = {Dynamic Prompt Learning via Policy Gradient for Semi-structured Mathematical
                  Reasoning},
  booktitle    = {{ICLR}},
  publisher    = {OpenReview.net},
  year         = {2023}
}

@inproceedings{DBLP:conf/acl/PasupatL15,
  author       = {Panupong Pasupat and
                  Percy Liang},
  title        = {Compositional Semantic Parsing on Semi-Structured Tables},
  booktitle    = {{ACL} {(1)}},
  pages        = {1470--1480},
  publisher    = {The Association for Computer Linguistics},
  year         = {2015}
}

@inproceedings{DBLP:conf/acl/ZhuLHWZLFC20,
  author       = {Fengbin Zhu and
                  Wenqiang Lei and
                  Youcheng Huang and
                  Chao Wang and
                  Shuo Zhang and
                  Jiancheng Lv and
                  Fuli Feng and
                  Tat{-}Seng Chua},
  title        = {{TAT-QA:} {A} Question Answering Benchmark on a Hybrid of Tabular
                  and Textual Content in Finance},
  booktitle    = {{ACL/IJCNLP} {(1)}},
  pages        = {3277--3287},
  publisher    = {Association for Computational Linguistics},
  year         = {2021}
}

@inproceedings{DBLP:conf/iclr/ChenWCZWLZW20,
  author       = {Wenhu Chen and
                  Hongmin Wang and
                  Jianshu Chen and
                  Yunkai Zhang and
                  Hong Wang and
                  Shiyang Li and
                  Xiyou Zhou and
                  William Yang Wang},
  title        = {TabFact: {A} Large-scale Dataset for Table-based Fact Verification},
  booktitle    = {{ICLR}},
  publisher    = {OpenReview.net},
  year         = {2020}
}

@inproceedings{DBLP:conf/acl/GuptaMNS20,
  author       = {Vivek Gupta and
                  Maitrey Mehta and
                  Pegah Nokhiz and
                  Vivek Srikumar},
  title        = {{INFOTABS:} Inference on Tables as Semi-structured Data},
  booktitle    = {{ACL}},
  pages        = {2309--2324},
  publisher    = {Association for Computational Linguistics},
  year         = {2020}
}

@inproceedings{DBLP:conf/acl/Cheng0WJG0HLZ22,
  author       = {Zhoujun Cheng and
                  Haoyu Dong and
                  Zhiruo Wang and
                  Ran Jia and
                  Jiaqi Guo and
                  Yan Gao and
                  Shi Han and
                  Jian{-}Guang Lou and
                  Dongmei Zhang},
  title        = {HiTab: {A} Hierarchical Table Dataset for Question Answering and Natural
                  Language Generation},
  booktitle    = {{ACL} {(1)}},
  pages        = {1094--1110},
  publisher    = {Association for Computational Linguistics},
  year         = {2022}
}

@inproceedings{DBLP:conf/nips/ZhaoFLTWLWY0ZLH24,
  author       = {Weichao Zhao and
                  Hao Feng and
                  Qi Liu and
                  Jingqun Tang and
                  Binghong Wu and
                  Lei Liao and
                  Shu Wei and
                  Yongjie Ye and
                  Hao Liu and
                  Wengang Zhou and
                  Houqiang Li and
                  Can Huang},
  title        = {TabPedia: Towards Comprehensive Visual Table Understanding with Concept
                  Synergy},
  booktitle    = {NeurIPS},
  year         = {2024}
}

@inproceedings{DBLP:conf/cvpr/ZhouGWZWCX25,
  author       = {Bangbang Zhou and
                  Zuan Gao and
                  Zixiao Wang and
                  Boqiang Zhang and
                  Yuxin Wang and
                  Zhineng Chen and
                  Hongtao Xie},
  title        = {SynTab-LLaVA: Enhancing Multimodal Table Understanding with Decoupled
                  Synthesis},
  booktitle    = {{CVPR}},
  pages        = {24796--24806},
  publisher    = {Computer Vision Foundation / {IEEE}},
  year         = {2025}
}

@article{DBLP:journals/corr/abs-2502-13923,
  author       = {Shuai Bai and
                  Keqin Chen and
                  Xuejing Liu and
                  Jialin Wang and
                  Wenbin Ge and
                  Sibo Song and
                  Kai Dang and
                  Peng Wang and
                  Shijie Wang and
                  Jun Tang and
                  Humen Zhong and
                  Yuanzhi Zhu and
                  Ming{-}Hsuan Yang and
                  Zhaohai Li and
                  Jianqiang Wan and
                  Pengfei Wang and
                  Wei Ding and
                  Zheren Fu and
                  Yiheng Xu and
                  Jiabo Ye and
                  Xi Zhang and
                  Tianbao Xie and
                  Zesen Cheng and
                  Hang Zhang and
                  Zhibo Yang and
                  Haiyang Xu and
                  Junyang Lin},
  title        = {Qwen2.5-VL Technical Report},
  journal      = {CoRR},
  volume       = {abs/2502.13923},
  year         = {2025}
}

@misc{gao2026relationaldatabasedistillationstructured,
      title={Relational Database Distillation: From Structured Tables to Condensed Graph Data}, 
      author={Xinyi Gao and Jingxi Zhang and Lijian Chen and Tong Chen and Lizhen Cui and Hongzhi Yin},
      year={2026},
      eprint={2510.06980},
      archivePrefix={arXiv},
      primaryClass={cs.DB},
      url={https://arxiv.org/abs/2510.06980}, 
}

@misc{yuan2026integratingvisioncentrictextunderstanding,
      title={Integrating Vision-Centric Text Understanding for Conversational Recommender Systems}, 
      author={Wei Yuan and Shutong Qiao and Tong Chen and Quoc Viet Hung Nguyen and Zi Huang and Hongzhi Yin},
      year={2026},
      eprint={2601.13505},
      archivePrefix={arXiv},
      primaryClass={cs.IR},
      url={https://arxiv.org/abs/2601.13505}, 
}

@misc{qiao2026textasvisionmeetssemanticids,
      title={When Text-as-Vision Meets Semantic IDs in Generative Recommendation: An Empirical Study}, 
      author={Shutong Qiao and Wei Yuan and Tong Chen and Xiangyu Zhao and Quoc Viet Hung Nguyen and Hongzhi Yin},
      year={2026},
      eprint={2601.14697},
      archivePrefix={arXiv},
      primaryClass={cs.IR},
      url={https://arxiv.org/abs/2601.14697}, 
}

@article{DBLP:journals/tkde/HungVTWYZ18,
  author       = {Nguyen Quoc Viet Hung and
                  Huynh Huu Viet and
                  Thanh Tam Nguyen and
                  Matthias Weidlich and
                  Hongzhi Yin and
                  Xiaofang Zhou},
  title        = {Computing Crowd Consensus with Partial Agreement},
  journal      = {{IEEE} Trans. Knowl. Data Eng.},
  volume       = {30},
  number       = {1},
  pages        = {1--14},
  year         = {2018}
}

@article{DBLP:journals/dase/YinQCYZLXSZ25,
  author       = {Hongzhi Yin and
                  Liang Qu and
                  Tong Chen and
                  Wei Yuan and
                  Ruiqi Zheng and
                  Jing Long and
                  Xin Xia and
                  Yuhui Shi and
                  Chengqi Zhang},
  title        = {On-Device Recommender Systems: {A} Comprehensive Survey},
  journal      = {Data Sci. Eng.},
  volume       = {10},
  number       = {4},
  pages        = {591--620},
  year         = {2025}
}

@inproceedings{DBLP:conf/sigir/RenYCWH021,
  author       = {Xuhui Ren and
                  Hongzhi Yin and
                  Tong Chen and
                  Hao Wang and
                  Zi Huang and
                  Kai Zheng},
  title        = {Learning to Ask Appropriate Questions in Conversational Recommendation},
  booktitle    = {{SIGIR}},
  pages        = {808--817},
  publisher    = {{ACM}},
  year         = {2021}
}

@article{DBLP:journals/tois/RenYCWHHZ20,
  author       = {Xuhui Ren and
                  Hongzhi Yin and
                  Tong Chen and
                  Hao Wang and
                  Nguyen Quoc Viet Hung and
                  Zi Huang and
                  Xiangliang Zhang},
  title        = {{CRSAL:} Conversational Recommender Systems with Adversarial Learning},
  journal      = {{ACM} Trans. Inf. Syst.},
  volume       = {38},
  number       = {4},
  pages        = {34:1--34:40},
  year         = {2020}
}

@misc{rcc_bunya_2024,
  author = {{The University of Queensland Research Computing Centre}},
  title  = {Bunya supercomputer},
  year   = {2024},
  address = {Brisbane, Queensland, Australia},
  howpublished = {\url{https://dx.doi.org/10.48610/wf6c-qy55}}
}
\bibliographystyle{iclr2026_conference}

\appendix
\section{Appendix}

\subsection{Reproducibility Statement}
\label{appendix:reproducibility}

We are committed to ensuring the reproducibility of our work. All source code, trained gating network checkpoints, and experiment configurations are provided at the following GitHub repository: \href{https://github.com/xiaobo-xing/TableDART}{https://github.com/xiaobo-xing/TableDART}. Below is a summary of the key components for reproduction.

\begin{itemize}
    \item \textbf{Model Architecture}: The overall architecture of TableDART is detailed in Section~\ref{sec:methodology}. Specific details of the embedding extraction pipeline and the MLP gating network are provided in Appendix~\ref{appendix:architecture-details}. The \textbf{primary} constituent models are TableGPT2-7B \citep{su2024tablegpt2largemultimodalmodel} and Ovis2-8B \citep{lu2024ovisstructuralembeddingalignment}, with Qwen2.5-VL-7B \citep{DBLP:journals/corr/abs-2502-13923} additionally employed to validate generalization. Google's Gemini 2.0 Flash model \citep{comanici2025gemini} implements the Fusion agent.
    
    \item \textbf{Training Procedure}: Our parameter-efficient training methodology for the gating network, including the resource-aware objective function, is described in Section~\ref{sec:gating_training}. Full implementation details are provided in Appendix~\ref{appendix:experimental_setup_details}.
    
    \item \textbf{Datasets and Preprocessing}: The datasets used for training and evaluation are described in Section~\ref{sec:experiments}. Our training mixture was constructed from five datasets: WTQ \citep{DBLP:conf/acl/PasupatL15}, TABMWP \citep{DBLP:conf/iclr/Lu0CWZRCK23}, TAT-QA \citep{DBLP:conf/acl/ZhuLHWZLFC20}, TabFact \citep{DBLP:conf/iclr/ChenWCZWLZW20}, and InfoTabs \citep{DBLP:conf/acl/GuptaMNS20}. We evaluate on these five datasets plus two zero-shot datasets: HiTab \citep{DBLP:conf/acl/Cheng0WJG0HLZ22} and FeTaQA \citep{DBLP:journals/tacl/NanHMLVZKSKTMRT22}. Our specific data construction protocol is detailed in Section~\ref{sec:implementation_details} and Appendix~\ref{appendix:experimental_setup_details}.
    
    \item \textbf{Hyperparameters and Computational Infrastructure}: A comprehensive list of all hyperparameters is provided in Table~\ref{tab:hyperparameters}. All experiments were conducted on a single NVIDIA H100 80GB GPU. These computational resources were provided by the Bunya supercomputer \citep{rcc_bunya_2024}, operated by The University of Queensland Research Computing Centre (RCC). Full details on the training pipeline, hyperparameters, and computational setup are located in Appendix~\ref{appendix:experimental_setup_details}.
\end{itemize}

\subsection{Large Language Models Usage}
Large Language Models were employed to fix the grammar issues in this paper. The authors remain fully responsible for all content.

\subsection{Model Architecture Details}
\label{appendix:architecture-details}

\subsubsection{Embedding Architecture and Computational Analysis}
\label{appendix:embedding-analysis}

The gating network requires fixed-dimensional feature representations to enable consistent routing decisions across variable-length inputs. We employ different pooling strategies optimized for each modality's characteristics:

\begin{itemize}
    \item \textbf{Table-as-Text Model Features}: 3,584 dimensions obtained through attention-masked mean pooling of TableGPT2-7B input embeddings. This approach weights each token by its attention importance, preventing padding tokens from diluting the semantic representation while preserving global table structure information essential for routing decisions.
    
    \item \textbf{Table-as-Image Model Features}: 6,144 dimensions derived from spatial-temporal pooling of Ovis2-8B visual tokenizer outputs. Mean pooling across spatial dimensions captures holistic visual layout characteristics rather than fine-grained spatial details, providing sufficient information for modality appropriateness assessment without overwhelming the routing mechanism with local visual features.
    
    \item \textbf{Question Embedding}: 384 dimensions from all-MiniLM-L6-v2 \citep{DBLP:conf/emnlp/ReimersG19} sentence transformer. This pre-trained model provides question-type agnostic semantic representations that complement the features from the single-modality models, enabling the gating network to learn routing patterns based on question semantics rather than model-specific encodings.
\end{itemize}

\textbf{Total Concatenated Dimension}: 3,584 + 6,144 + 384 = 10,112 dimensions. This unified multi-modal representation serves as the input to the MLP gating network, which learns to map these heterogeneous features to processing path selection probabilities.

\textbf{Parameter Analysis}: The embedding extraction phase utilizes only a small fraction of each model's total parameters: (1) The Table-as-Text model requires 545M parameters from the input embedding layer (152K vocab × 3,584 dimensions), representing 7.15\% of TableGPT2-7B's 7.62B parameters; (2) The Table-as-Image model requires 682M parameters from the Vision Transformer component aimv2-huge-patch14-448 for visual feature extraction, representing 7.63\% of Ovis2-8B's 8.94B parameters. This computational efficiency enables lightweight routing decisions while preserving rich semantic representations from the pre-trained models.

\subsubsection{Gating Network Architecture Specification}
\label{appendix:gating-architecture}

Following standard practice in dynamic routing and mixture-of-experts systems in the prior works \citep{DBLP:conf/iclr/LepikhinLXCFHKS21,DBLP:conf/cvpr/ZhuZLWLWD22,DBLP:conf/cvpr/XueM23}, the MLP gating network implements a standard two-layer architecture designed for efficient and effective routing. The detailed forward pass is defined as:
\begin{align}
\mathbf{h} &= \text{ReLU}(\mathbf{W}_1 \mathbf{x} + \mathbf{b}_1) \label{eq:gate-layer1} \\
\mathbf{h}' &= \text{Dropout}(\mathbf{h}, p=0.1) \label{eq:gate-dropout} \\
\mathbf{z} &= \mathbf{W}_2 \mathbf{h}' + \mathbf{b}_2 \label{eq:gate-layer2}
\end{align}
where $\mathbf{x} \in \mathbb{R}^{10112}$ is the concatenated multi-modal input features, and $\mathbf{z} \in \mathbb{R}^3$ represents the final expert selection logits. During inference, these logits are converted to probabilities via a Softmax function to select the optimal expert.

\textbf{Implementation Details.} The hidden dimension is set to 256, a choice empirically found to balance representational capacity for learning complex routing patterns with computational efficiency. Dropout with a probability of $p=0.1$ is applied after the hidden layer to prevent overfitting. We use the ReLU activation function for its computational efficiency and stable gradient properties. All weight matrices are initialized using Xavier uniform initialization \citep{DBLP:journals/jmlr/GlorotB10}, and bias vectors are initialized to zero to ensure stable training dynamics.

\textbf{Parameter Count.} The lightweight nature of our gating network is evident in its parameter count, detailed in Table~\ref{tab:gating-parameters}. The total of \textbf{2.59M} trainable parameters is negligible compared to the billions of parameters in the frozen LLM/MLLM models, underscoring the high parameter efficiency of our TableDART framework.

% Table 
\begin{table}[htbp]
\centering
\small
\begin{tabular}{lcc}
\toprule
\textbf{Component} & \textbf{Computation} & \textbf{Parameters} \\
\midrule
\multicolumn{3}{l}{\textbf{First Layer}} \\[0.5ex]
Weight matrix $\mathbf{W}_1$ & $10{,}112 \times 256$ & $2{,}588{,}672$ \\
Bias vector $\mathbf{b}_1$ & $256$ & $256$ \\
\textit{First layer subtotal} & & $2{,}588{,}928$ \\[1ex]
\multicolumn{3}{l}{\textbf{Second Layer}} \\[0.5ex]
Weight matrix $\mathbf{W}_2$ & $256 \times 3$ & $768$ \\
Bias vector $\mathbf{b}_2$ & $3$ & $3$ \\
\textit{Second layer subtotal} & & $771$ \\[1ex]
\midrule
\textbf{Total Network Parameters} & & 
\textbf{2{,}589{,}699} $\approx$ \textbf{2.59M} \\
\bottomrule
\end{tabular}
\caption{Detailed parameter breakdown of the MLP gating network.}
\label{tab:gating-parameters}
\end{table}

% Fusion Expert Detailed Design Section
\subsection{Fusion Path Detailed Design}
\label{appendix:fusion-design}

\begin{figure}[ht]
    \centering
    \includegraphics[width=0.9\columnwidth]{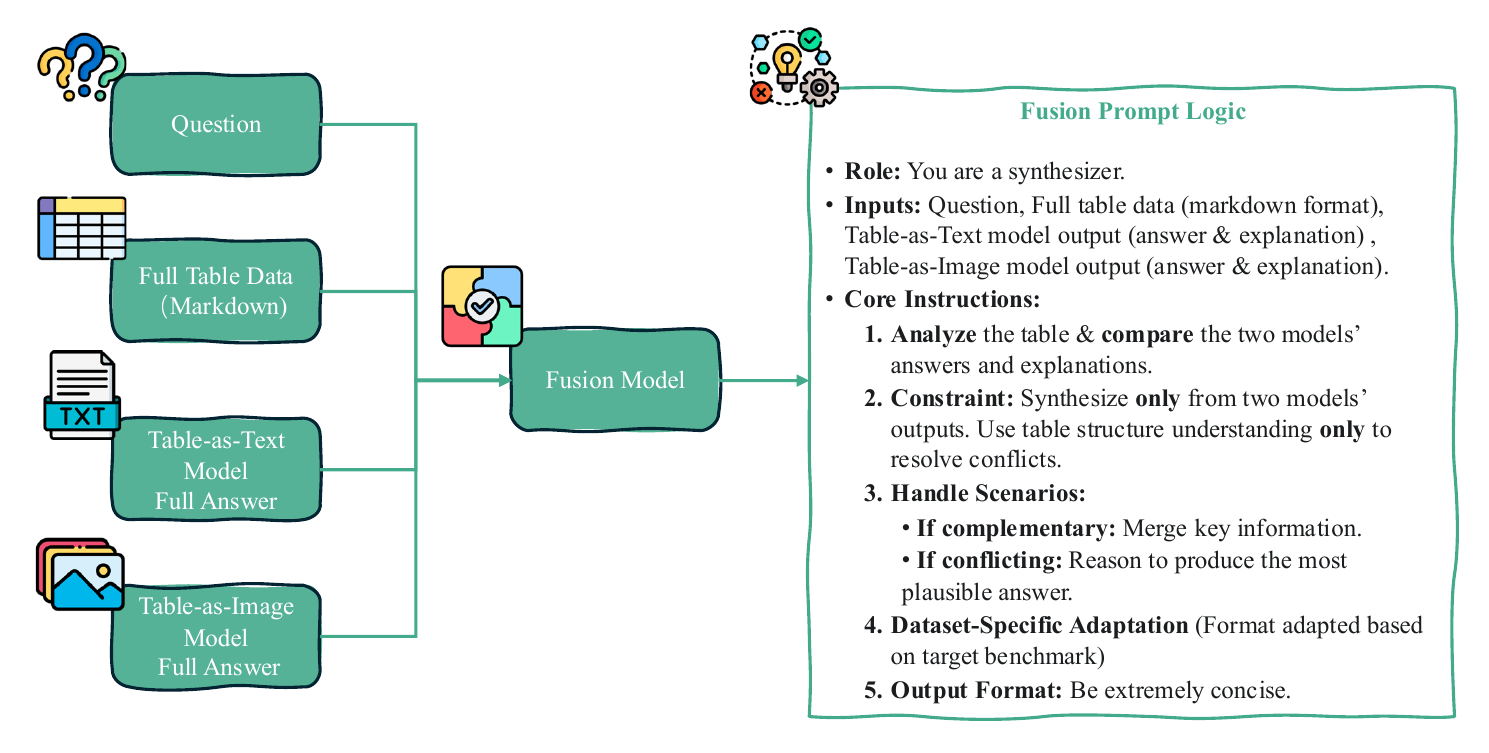}
     \caption{The data flow and logical structure of the prompt for the Fusion path's LLM agent. This structure guides the synthesis of a final answer from the outputs of the single-modality models.}
    \label{fig:fusion-prompt}
\end{figure}

This section provides the detailed design of the Fusion path, which serves as the sophisticated synthesis mechanism within the TableDART framework. As described in the main text, this component is implemented as an LLM agent that interfaces with Google’s Gemini 2.0 Flash model via REST API calls. For each instance, the LLM agent receives four key inputs: (1) the original question, (2) the complete table data in a structured markdown format, (3) the output from the Table-as-Text model containing its answer and explanation, and (4) the output from the Table-as-Image model containing its answer and explanation. These inputs are formatted into a prompt whose logical structure is illustrated in Figure~\ref{fig:fusion-prompt}. To ensure output format compatibility across benchmarks, the prompt is dynamically adapted based on the target dataset. As detailed in Table~\ref{tab:dataset-adaptations}, specific instructions are appended to the prompt to meet the unique formatting requirements of each benchmark, such as requiring JSON outputs for TabFact or full-sentence responses for FeTaQA.

\begin{table}[htbp]
\centering
\small
\begin{tabular}{l p{9cm}}
\toprule
\textbf{Dataset(s)} & \textbf{Additional Instruction} \\
\midrule
TabFact & Generate JSON response with ``answer" field containing [``True"] or [``False"] \\
InfoTabs & Generate JSON response with ``answer" field: [``Entail"], [``Contradict"], or [``Neutral"] \\
TabMWP & Output numeric answers without units when applicable \\
FeTaQA & Provide complete sentence responses, not keywords or phrases \\
WTQ, HiTab, TAT-QA & Standard JSON format with concise, direct answers \\
\bottomrule
\end{tabular}
\caption{Dataset-specific prompt adaptations for the Fusion path's LLM agent.}
\label{tab:dataset-adaptations}
\end{table}

\subsection{Experimental Setup and Implementation Details}
\label{appendix:experimental_setup_details}

\subsubsection{Datasets and Metrics}
Our experimental evaluation is conducted on seven diverse benchmarks across two primary tasks: Table Question Answering (TQA) and Table Fact Verification (TFV). We follow the established protocol of previous works \citep{DBLP:conf/acl/ZhengFSS0J024,liu2025hippoenhancingtableunderstanding} for evaluation metrics, using Accuracy for most tasks and BLEU score for the generative FeTaQA benchmark. Table~\ref{tab:data-statistics} provides a detailed breakdown of these datasets, including the number of instances used for our training mixture and the full size of the official test sets.

\begin{table}[h]
\centering
\small
\begin{tabular}{llccc}
\toprule
\textbf{Task Type} & \textbf{Dataset} & \textbf{Train} & \textbf{Validation} & \textbf{Test} \\
\midrule
\multirow{5}{*}{\makecell[l]{Table Question\\Answering (TQA)}} 
& TABMWP & 2,000 & 300 & 7,686 \\
& WTQ & 2,000 & 300 & 4,344 \\
& HiTab & --- & --- & 1,586 \\
& TAT-QA & 2,000 & 300 & 772 \\
& FeTaQA & --- & --- & 2,003 \\
\midrule
\multirow{2}{*}{\makecell[l]{Table Fact\\Verification (TFV)}} 
& TabFact & 2,000 & 300 & 6,845 \\
& InfoTabs & 2,000 & 300 & 5,400 \\
\midrule
\multicolumn{2}{l}{\textbf{Total}} & \textbf{10,000} & \textbf{1,500} & \textbf{28,636} \\
\bottomrule
\end{tabular}
\caption{Statistics of the datasets used for our experiments. The `Test' column reflects the official benchmark size, while the `Train' and `Validation' columns indicate the number of instances we sampled from the official training data to construct our training mixture, as detailed in Section~\ref{sec:implementation_details}.}
\label{tab:data-statistics}
\end{table}

\subsubsection{Comparative Models and Rationale}
\label{appendix:models_relationale}
We compare TableDART against a comprehensive set of baselines, strategically selected to validate our core contributions across several axes:
\begin{itemize}[itemsep=2pt, topsep=4pt]
    \item \textbf{Constituent Models:} To demonstrate that TableDART's performance stems from its dynamic framework, we evaluate its core constituent models as standalone baselines. These include: (1) the Table-as-Text model, TableGPT2-7B \citep{su2024tablegpt2largemultimodalmodel}; (2) the primary Table-as-Image model, Ovis2-8B \citep{lu2024ovisstructuralembeddingalignment}; (3) the additional visual expert used for generalization, Qwen2.5-VL-7B \citep{DBLP:journals/corr/abs-2502-13923}; and (4) the backbone of our Fusion agent, Google Gemini 2.0 Flash \citep{comanici2025gemini}. We selected these backbone models because they are recent and strong representatives of their respective paradigms. Evaluating the single-modality experts in isolation establishes a performance baseline. Specifically, including Qwen2.5-VL-7B enables us to test the framework's generalization capabilities across different visual backbones, while comparing against Gemini as a standalone MLLM verifies that our performance lift arises from the routing mechanism rather than only the capacity of the Fusion agent alone.

    \item \textbf{MLLM-based Baselines:} To highlight the superiority of our dynamic adaptive routing paradigm over static, one-size-fits-all approaches, we compare against two key models. First, HIPPO \citep{liu2025hippoenhancingtableunderstanding}, a recent model representing the MLLM paradigm that jointly processes both text and image representations for all inputs. Second, Google Gemini 2.0 Flash \citep{comanici2025gemini}. Since Gemini serves as the backbone for the LLM agent in our Fusion path, including it as a standalone MLLM baseline allows us to demonstrate that the observed improvements stem from our dynamic routing framework itself, rather than solely from the capacity of the Fusion agent's underlying model.
    \item \textbf{Broader Competitive Landscape:} To position TableDART within the broader field, we benchmark it against an extensive suite of single-modality baselines in the 7-8B parameter range. Our selection of Table-as-Text models includes generalist LLMs like Llama-2-7B \citep{touvron2023llama} and Llama3-Instruct-8B \citep{dubey2024llama}, as well as the specialized TableLlama-7B \citep{DBLP:conf/naacl/ZhangYL024}. The set of Table-as-Image MLLMs is equally comprehensive, featuring MiniGPT-4-7B \citep{DBLP:conf/iclr/Zhu0SLE24}, mPLUG-Owl-7B \citep{ye2023mplug}, mPLUG-Owl2-7B \citep{ye2024mplug}, LLaVA v1.5-7B \citep{DBLP:conf/cvpr/LiuLLL24}, Table-LLaVA-7B \citep{DBLP:conf/acl/ZhengFSS0J024}, Qwen-VL-7B \citep{bai2023qwen}, InternLM-XComposer2-7B \citep{zhang2023internlm}, Monkey-7B \citep{DBLP:conf/cvpr/LiYLMZYSLB24}, TabPedia-7B \citep{DBLP:conf/nips/ZhaoFLTWLWY0ZLH24}, SynTab-LLaVA-7B \citep{DBLP:conf/cvpr/ZhouGWZWCX25}, Qwen2.5-VL-7B \citep{DBLP:journals/corr/abs-2502-13923} and MiniCPM-V-2.6-8B \citep{yao2024minicpm}.
\end{itemize}

For a comprehensive and fair comparison, we report several baseline results directly from their original publications. 
Specifically, the results for the following models are adopted from the HIPPO paper \citep{liu2025hippoenhancingtableunderstanding}: Llama-2-7B, Llama3-Instruct-8B, TableLlama-7B, MiniGPT-4-7B, mPLUG-Owl-7B, mPLUG-Owl2-7B, LLaVA v1.5, Table-LLaVA-7B, MiniCPM-V-2.6-8B, Qwen-VL-7B, InternLM-XComposer2-7B, Monkey-7B, and the HIPPO model itself.
Furthermore, the results for TableGPT2-7B are sourced from its proposal paper \citep{su2024tablegpt2largemultimodalmodel}, and the results for SynTab-LLaVA are sourced from \citet{DBLP:conf/cvpr/ZhouGWZWCX25}.
We also adopt the results for TabPedia and Ovis2-8B (except TABMWP dataset) from \citet{jiang2025multimodaltabularreasoningprivileged}. 
All other model results not mentioned above were generated by our own experimental runs.

\subsubsection{Training and Implementation Details}
\label{appendix:training-implementation}
\paragraph{Training Data Construction.}
Our training set is a 10,000-sample mixture, constructed by randomly sampling 2,000 instances from five datasets: TABMWP, WTQ, TAT-QA, TabFact, and InfoTabs \citep{DBLP:conf/iclr/Lu0CWZRCK23, DBLP:conf/acl/PasupatL15, DBLP:conf/acl/ZhuLHWZLFC20, DBLP:conf/iclr/ChenWCZWLZW20, DBLP:conf/acl/GuptaMNS20}. Following the protocol of HIPPO \citep{liu2025hippoenhancingtableunderstanding}, FeTaQA and HiTab are excluded from training due to their non-accuracy-based evaluation metric (BLEU) and challenges in serializing complex hierarchical structures, respectively \citep{DBLP:journals/tacl/NanHMLVZKSKTMRT22, DBLP:conf/acl/Cheng0WJG0HLZ22}. An additional 15\% validation split (1,500 samples) is generated using the same random sampling procedure with a fixed seed to ensure reproducibility.

\paragraph{Processing Path Cost Measurement Protocol.}
The resource cost vector $\mathbf{c}$ used in our resource-aware objective is derived from empirical measurements to provide a realistic estimate of each path's computational expense. We measured these costs on a representative testbed of 70 samples (10 from each benchmark), averaging over 10 timed runs following 5 warm-up iterations. Our final cost metric offers a holistic balance of latency (seconds per instance) and throughput (tokens per second), defined as: $\text{Cost} = 0.5 \times (\text{Avg Latency}) + 0.5 \times (1.0 / \text{Avg TPS})$. For the Fusion path, its latency assumes parallel execution of its base models plus a 0.3s API average overhead, reflecting a realistic deployment scenario. The resulting empirically derived cost values are summarized in Table~\ref{tab:cost_calculation}.

\begin{table}[h]
\centering
\small
\caption{Breakdown of the processing path cost calculation. The Fusion path's latency assumes parallel execution of the two single-modality models (taking the max latency) plus a measured API overhead. Its TPS is consequently bottlenecked by the slower Table-as-Image model.}
\label{tab:cost_calculation}
\begin{tabular}{l ccc}
\toprule
\textbf{Processing Path} & \textbf{Avg. Latency (s)} & \textbf{Avg. TPS} & \textbf{Final Cost ($\mathbf{c}$)} \\
\midrule
Text-only & 1.445 & 44.19 & 0.73 \\
Image-only & 1.559 & 18.78 & 0.81 \\
Fusion & 1.859 & 18.78 & 0.96 \\
\bottomrule
\end{tabular}
\end{table}

\paragraph{Hyperparameter Configuration.}
Table~\ref{tab:hyperparameters} summarizes the complete hyperparameter settings used for training the gating network. The resource loss weight $\lambda = 0.15$ is selected based on extensive ablation studies (Section~\ref{sec:ablation_mechanisms}), which demonstrate optimal performance-efficiency trade-offs at this value. These hyperparameters are selected through preliminary experiments to balance training stability and convergence speed. The target temperature $\tau = 0.3$ creates sufficiently sharp target distributions while preventing degenerate solutions, while the gate temperature $\tau_g = 1.0$ maintains standard softmax behavior during training.

\begin{table}[h]
\centering
\small
\begin{tabular}{lc}
\toprule
\textbf{Parameter} & \textbf{Value} \\
\midrule
Learning Rate & 1e-4 \\
Batch Size & 8 \\
Gradient Accumulation Steps & 4 \\
Effective Batch Size & 32 \\
Weight Decay & 0.01 \\
Gradient Clipping (Max Norm) & 1.0 \\
Hidden Dimension & 256 \\
Dropout Rate & 0.1 \\
Target Temperature ($\tau$) & 0.3 \\
Gate Temperature ($\tau_g$) & 1.0 \\
Resource Loss Weight ($\lambda$) & 0.15 \\
LR Warmup Ratio & 0.05 \\
Training Epochs & 1 \\
\bottomrule
\end{tabular}
\caption{Complete hyperparameter configuration for TableDART training.}
\label{tab:hyperparameters}
\end{table}

\paragraph{Training Pipeline.}
Expert model parameters remain frozen throughout training, with only the lightweight 2-layer MLP gating network being optimized. This parameter-efficient approach dramatically reduces computational requirements compared to joint fine-tuning alternatives. The learning rate follows a cosine annealing schedule with 5\% linear warmup, starting from zero and reaching the maximum learning rate of 1e-4 after the warmup period. We employ the AdamW optimizer with weight decay 0.01 to prevent overfitting of the gating network. Gradient accumulation over 4 steps achieves an effective batch size of 32 while maintaining memory efficiency on single-GPU training. Gradient clipping with maximum norm 1.0 ensures training stability, particularly important given the dynamic target generation process.

\paragraph{Computational Configuration and Convergence.}
Training is conducted on a single NVIDIA H100 80GB GPU with mixed precision optimization. Expert models utilize bfloat16 precision to reduce memory consumption, while the gating network maintains float32 precision for numerical stability during gradient computation. The complete training process requires approximately 13.5 hours to complete one epoch. Training converges rapidly within this single epoch, which we empirically determined to be optimal to prevent overfitting. The model is evaluated on the held-out validation set, and the checkpoint with the highest validation accuracy is saved as the best model for inference.

\subsection{Efficiency Benchmark Protocol}
\label{appendix:efficiency_protocol}

This section provides a detailed account of the protocol used for the efficiency analysis presented in Section~\ref{sec:main_results}. We outline the methodology for data sampling, performance measurement under a parallel assumption, and the precise definitions for all reported metrics to ensure full reproducibility.

\subsubsection{Benchmark Setup and Data Sampling}
To create a representative and manageable testbed, we constructed an evaluation set via stratified random sampling from the seven benchmark test sets. We randomly sampled a balanced set of 50 instances from each dataset, resulting in a comprehensive benchmark suite of 350 unique samples. To ensure statistical stability, the entire measurement process was repeated three times with different random seeds, and all reported metrics are the average across these independent runs. All benchmarks were executed on a single NVIDIA H100 80GB GPU.

\subsubsection{Measurement Protocol and Parallel Assumption}
Our measurement protocol is designed to simulate a realistic parallel-processing deployment scenario. For each sample, the total inference latency is calculated by summing the durations of three sequential phases, with parallelism applied within phases where appropriate.

\begin{itemize}[itemsep=2pt, topsep=4pt]
    \item \textbf{Phase 1: Parallel Feature Extraction.} The framework concurrently extracts embeddings from the input question, the serialized text table, and the table image. The duration of this phase is determined by the maximum latency among these three parallel operations: 
    $T_{\text{phase1}} = \max(T_{\text{text\_embed}}, T_{\text{vision\_embed}}, T_{\text{question\_embed}})$.

    \item \textbf{Phase 2: Gating and Routing.} The concatenated embeddings are fed into the lightweight gating network to yield a routing decision. This phase is only applicable to TableDART; for the Non-Adaptive Fusion baseline, its duration is zero.

    \item \textbf{Phase 3: Generation and Fusion.} The execution path depends on the routing decision:
    \begin{itemize}
        \item \textbf{For TableDART (Unimodal Path):} If the Text-only or Image-only path is selected, the duration is simply the generation time of the chosen model ($T_{\text{text\_gen}}$ or $T_{\text{vision\_gen}}$).
        \item \textbf{For TableDART (Fusion Path) and the Non-Adaptive Fusion baseline:} Both the Table-as-Text and Table-as-Image models perform generation in parallel, followed by a call to the Fusion API. The duration is the maximum of the two generation times plus the API call latency: 
        $T_{\text{phase3}} = \max(T_{\text{text\_gen}}, T_{\text{image\_gen}}) + T_{\text{fusion\_api}}$.
    \end{itemize}
\end{itemize}
The total latency for each sample is the sum of these three phases: $L_{\text{parallel}} = T_{\text{phase1}} + T_{\text{phase2}} + T_{\text{phase3}}$.

\subsubsection{Metric Definitions}
We use the following two primary metrics to report efficiency:

\begin{itemize}[itemsep=2pt, topsep=4pt]
    \item \textbf{Latency (s):} The total time in seconds to process a single sample, calculated as $L_{\text{parallel}}$ under the parallel assumption described above. This is our primary metric for comparing the end-to-end speed of different frameworks. Lower values are better.
    
    \item \textbf{Tokens per Second (TPS):} A measure of throughput, calculated by dividing the number of generated output tokens by the total parallel latency. The token count is estimated using the Text Expert's tokenizer. Higher values are better.
\end{itemize}

\subsection{Detailed Architectural Analysis}
\label{sec:appendix_detailed_analysis}

\begin{figure*}[t!]
    \centering
    \includegraphics[width=\textwidth]{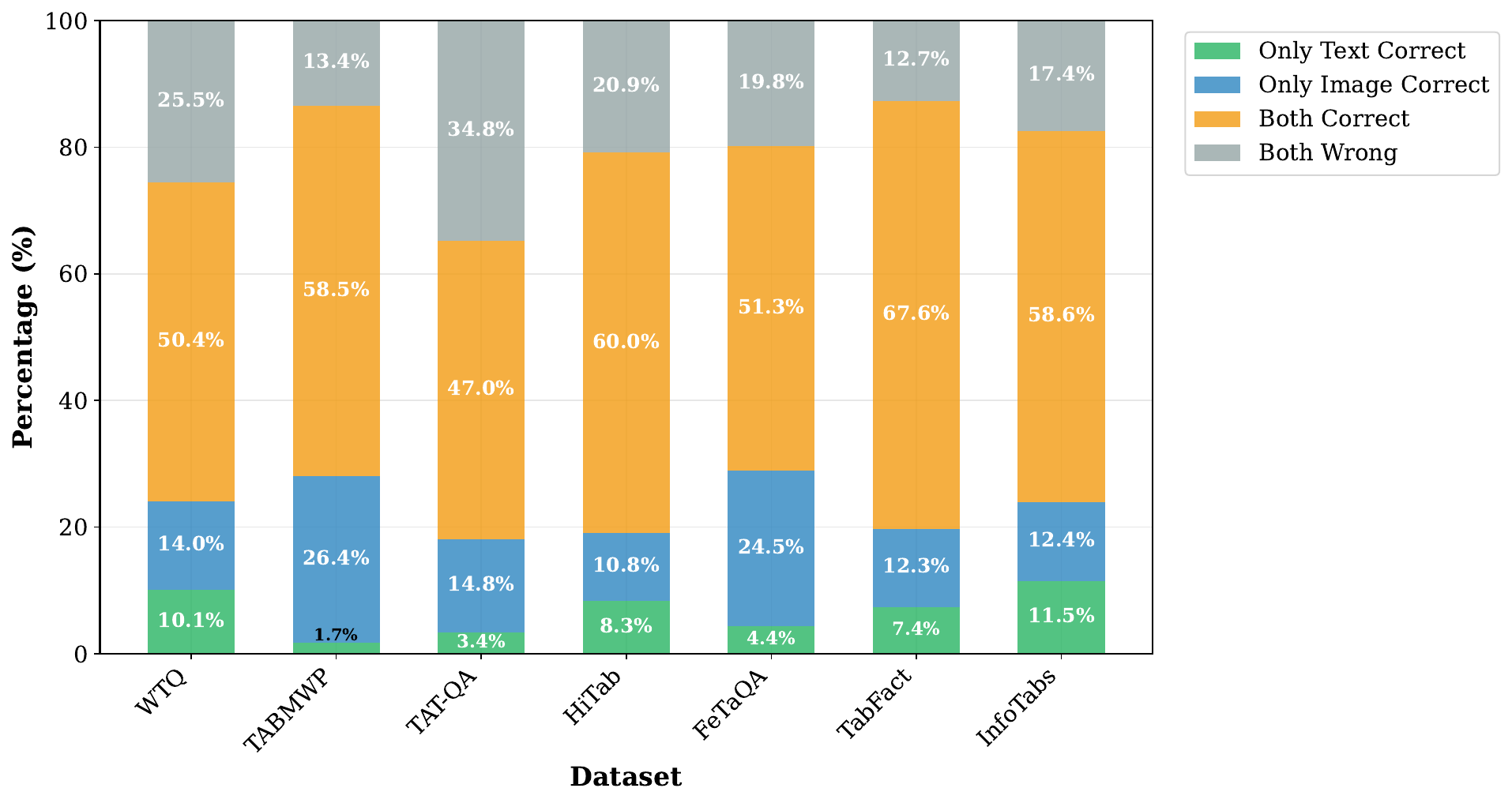}
    \caption{Per-dataset breakdown of performance overlap between the single-modality models. The stacked bars show the percentage of instances solved by only the Table-as-Text model, only the Table-as-Image model, both, or neither. The distribution varies significantly across benchmarks, highlighting the need for adaptive routing.}
    \label{fig:complementarity_breakdown}
\end{figure*}

\begin{figure*}[t!]
    \centering
    \includegraphics[width=\textwidth]{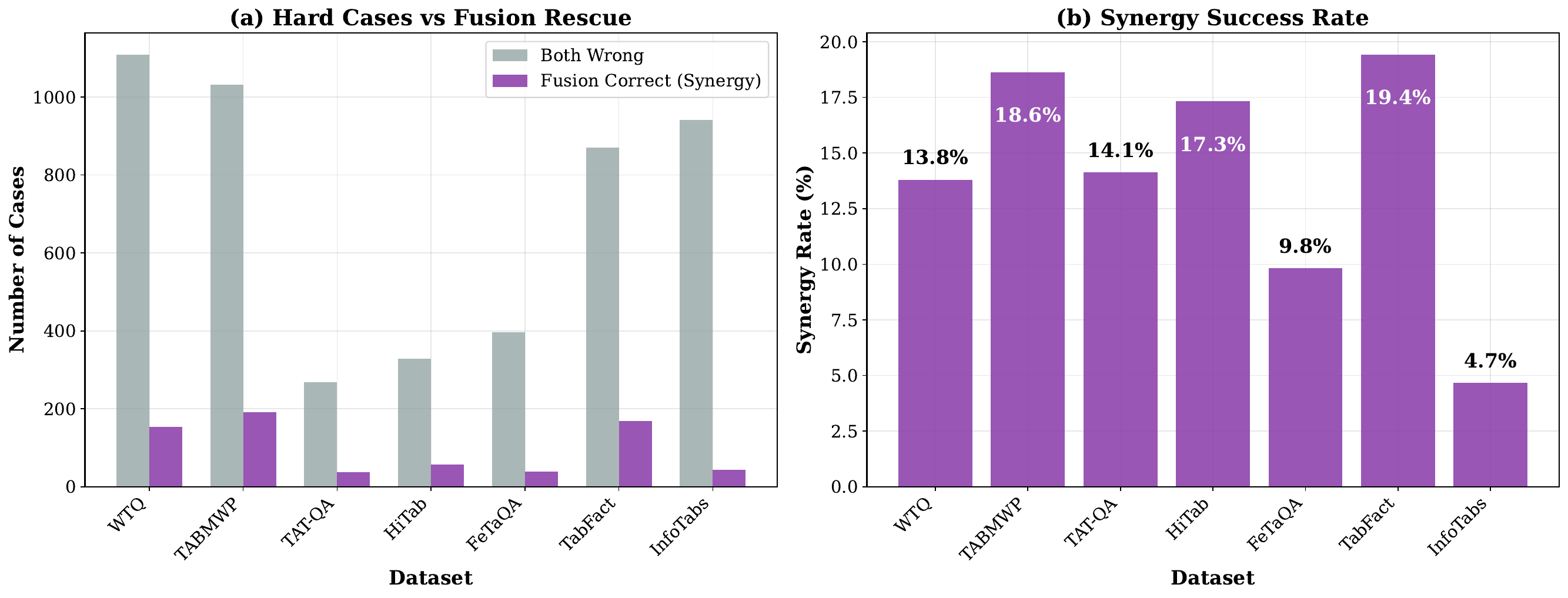}
    \caption{Detailed analysis of Fusion Path Synergy. (a) The absolute number of hard cases (where both base models failed) and the number of those cases successfully rescued by the Fusion path (synergy). (b) The Synergy Success Rate (Fusion Path Correct / Both Base Models Wrong) for each dataset demonstrates the consistent effectiveness of the Fusion path.}
    \label{fig:synergy_breakdown}
\end{figure*}

This section provides detailed visualizations to supplement the summary analysis in Section~\ref{sec:analysis_architecture}. Figure~\ref{fig:complementarity_breakdown} offers a full, per-dataset breakdown of the performance overlap between the single-modality models. Figure~\ref{fig:synergy_breakdown} provides a more detailed view of the synergy analysis, showing both the absolute number of hard cases and the corresponding synergy success rates for each benchmark. These detailed charts serve as the underlying evidence for the aggregate statistics and trends discussed in the main paper.

\section{Further In-depth Analysis}

\subsection{Full Gating Decision Analysis on All Datasets}
\label{sec:appendix_full_charts}

To supplement the analysis in the main paper, this section provides the complete visualizations of the processing path selection distribution for all seven benchmarks. Figure~\ref{fig:main_body_chart} in the main text highlights two representative datasets, while Figure~\ref{fig:appendix_chart} below shows the results on the remaining five.

The trends observed here are consistent with our main findings. For instance, on the math-heavy \textbf{TABMWP} dataset, the policy heavily favors the Image-only path. On the visually complex \textbf{HiTab} benchmark, the routing policy maintains a significant reliance on both the Image-only and Fusion paths across all $\lambda$ values. Finally, the generative \textbf{FeTaQA} task shows a strong preference for the Fusion path when performance is prioritized. These varied, dataset-specific behaviors further validate that TableDART successfully learns to adapt its strategy to the underlying characteristics of the data.

\begin{figure*}[t!]
    \centering
    \includegraphics[width=\textwidth]{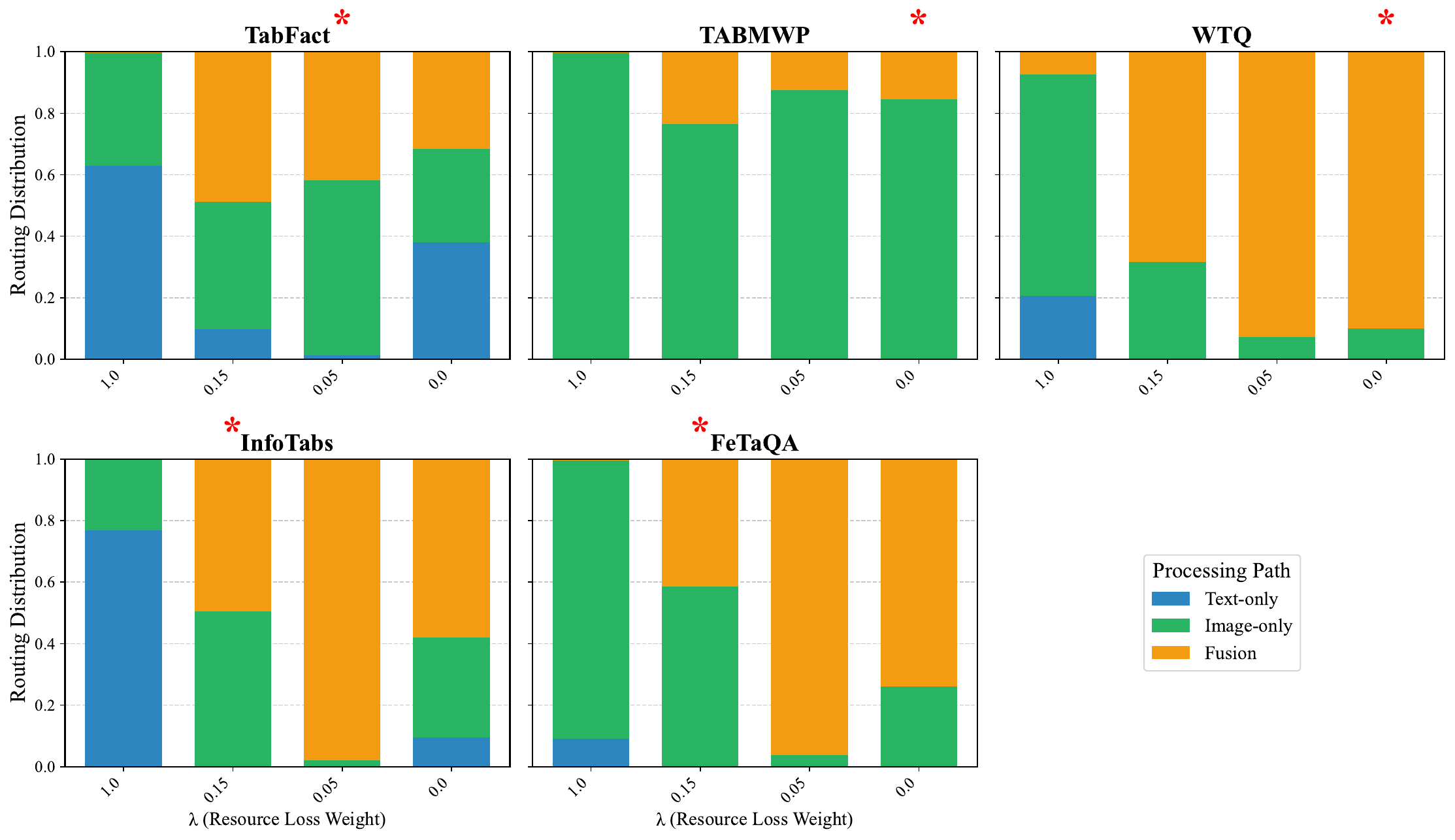}
    \caption{Processing path selection distribution vs. the resource loss weight ($\lambda$) on the five additional benchmarks. A red star (*) marks the best-performing configuration for each dataset. The data demonstrates consistent adaptive behavior across these diverse tasks.}
    \label{fig:appendix_chart}
\end{figure*}

\subsection{Analysis of the Learned Routing Policy}
\label{sec:analysis_gating}

Our final in-depth analysis dissects the sophisticated nature of the learned routing policy, revealing a crucial trade-off between aligning with a simple heuristic and achieving globally-optimal task performance. To investigate this, we adopt a behavioral analysis approach, a common technique in policy analysis and explainable AI \citep{DBLP:journals/csur/GuidottiMRTGP19}, by defining a Heuristic Alignment Score. This metric measures the percentage of times TableDART's routing policy matches a simple, greedy heuristic: always choose the most cost-effective processing path (first Text-only, then Image-only, then Fusion) that is known to solve a given problem correctly. The procedure for calculating this metric is detailed in Algorithm~\ref{alg:heuristic_alignment}. We then plot this alignment score against the final average task performance for each of our $\lambda$ configurations.

\begin{algorithm}[h!]
\caption{Heuristic Alignment Score Calculation}
\label{alg:heuristic_alignment}
\begin{algorithmic}[1]
\State \textbf{Input:} Set of test samples $S$, Table-as-Text model results $R_{Text}$, Table-as-Image model results $R_{Image}$, TableDART's routing decisions $D_{TableDART}$.
\State \textbf{Output:} Heuristic Alignment Score $S_{align}$.
\State
\State $N_{aligned\_routes} \gets 0$
\State $N_{total} \gets |S|$
\State
\For{each sample $s_i \in S$}
    \State $is\_text\_correct \gets R_{Text}(s_i) \text{ is correct}$
    \State $is\_image\_correct \gets R_{Image}(s_i) \text{ is correct}$
    \State $TableDART\_choice \gets D_{TableDART}(s_i)$
    \State
    \State \Comment{Define the simple, greedy heuristic action}
    \If{$is\_text\_correct$}
        \State $heuristic\_choice \gets \text{``Text-only"}$
    \ElsIf{$is\_image\_correct$}
        \State $heuristic\_choice \gets \text{``Image-only"}$
    \Else
        \State $heuristic\_choice \gets \text{``Fusion"}$
    \EndIf
    \State
    \If{$TableDART\_choice = heuristic\_choice$}
        \State $N_{aligned\_routes} \gets N_{aligned\_routes} + 1$
    \EndIf
\EndFor
\State
\State $S_{align} \gets (N_{aligned\_routes} / N_{total}) \times 100$
\State \textbf{return} $S_{align}$
\end{algorithmic}
\end{algorithm}

\begin{figure}[t!]
    \centering
    \includegraphics[width=0.80\columnwidth]{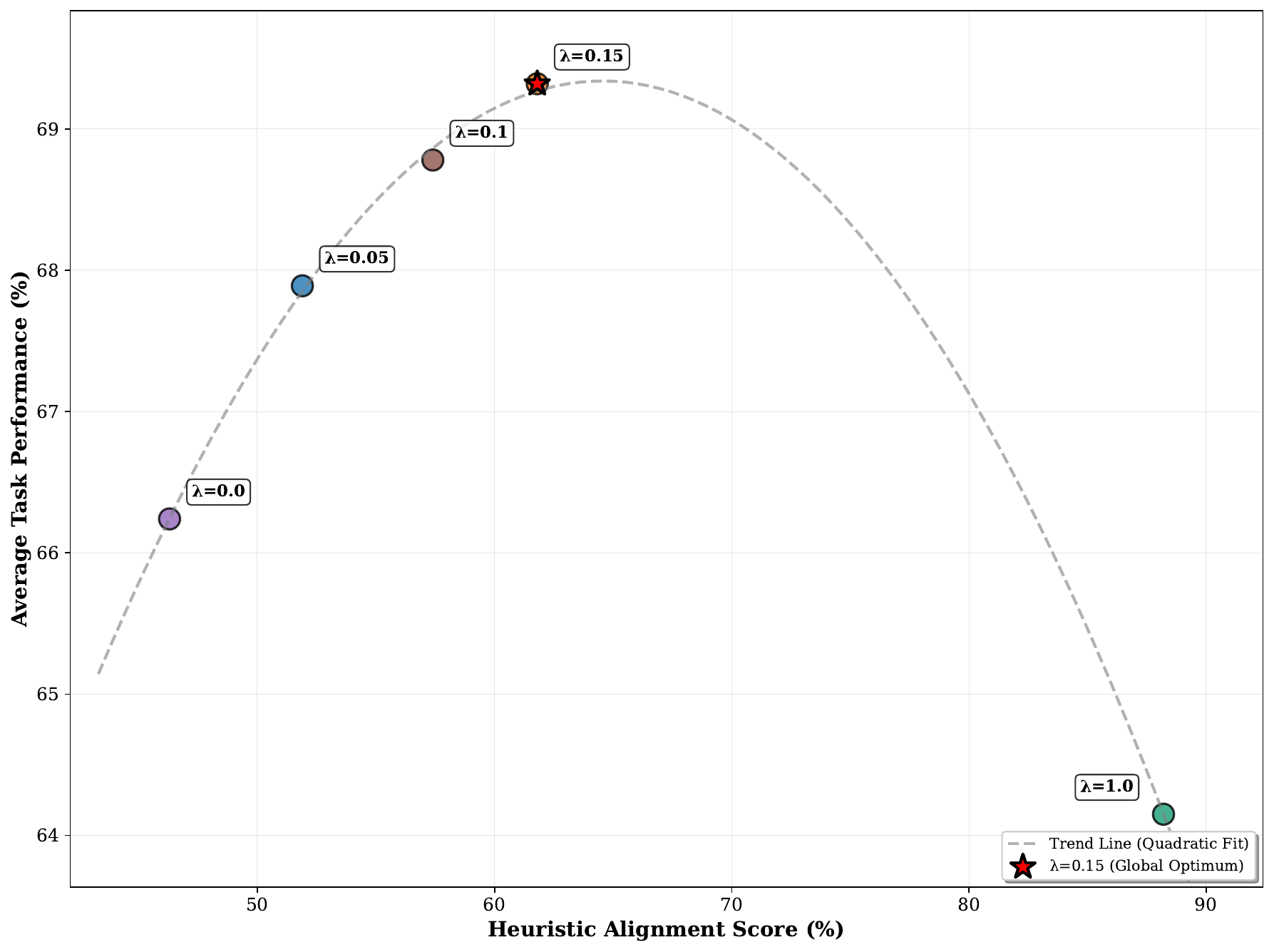}
    \caption{The trade-off between aligning with a simple greedy heuristic (x-axis) and achieving globally-optimal task performance (y-axis). The curve shows that strictly following the greedy heuristic can hurt final performance, and TableDART's best configuration ($\lambda=0.15$) learns a superior, non-greedy policy.}
    \label{fig:heuristic_vs_performance}
\end{figure}

The result, shown in Figure~\ref{fig:heuristic_vs_performance}, illustrates a clear non-linear relationship, resembling a Pareto frontier. We observe that a policy laser-focused on maximizing heuristic alignment (e.g., at $\lambda=1.0$, the rightmost point) adheres most closely to the simple greedy rule but yields a suboptimal overall task performance. This demonstrates that a greedy, locally-focused strategy can be detrimental to the global objective.

Crucially, our chosen configuration, $\lambda=0.15$, resides at the apex of this performance curve. Its modest Heuristic Alignment Score indicates that it has learned a more sophisticated, non-greedy policy. It quantitatively proves that the globally optimal strategy, learned via end-to-end training, involves strategically investing in more computationally expensive processing paths, even when a cheaper option is technically viable. This finding validates that TableDART learns a truly effective, globally-optimized routing policy that transcends simple heuristics.

\subsection{Analysis of Per-Dataset Routing Policies}
\label{sec:appendix_adaptive_behavior}

\begin{figure}[t!]
    \centering
    \includegraphics[width=\columnwidth]{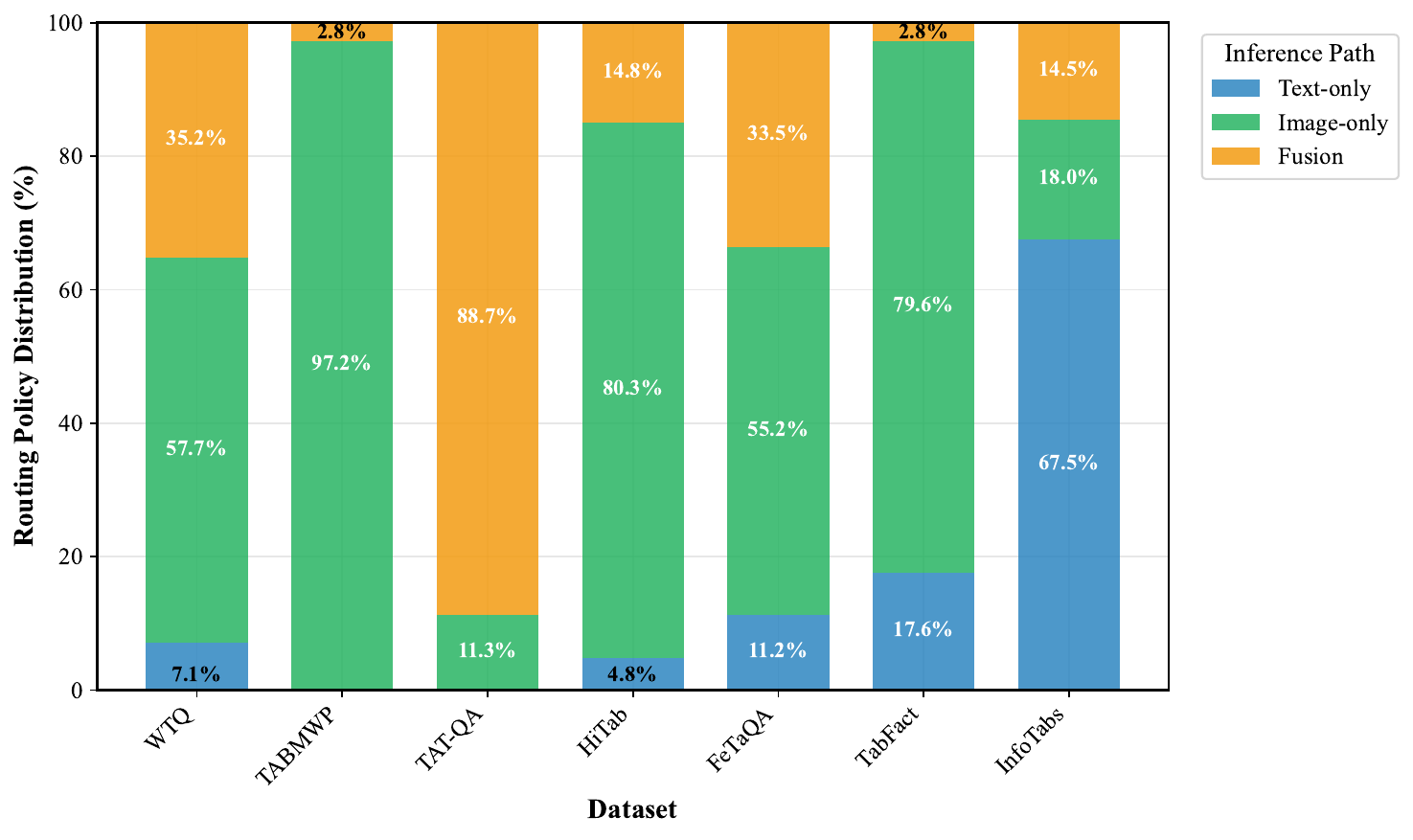} 
    \caption{The learned routing policy distribution for each benchmark using the final TableDART model ($\lambda=0.15$). The distinct strategies across datasets provide strong evidence of the framework's adaptive nature.}
    \label{fig:routing_distribution}
\end{figure}

To provide quantitative evidence of TableDART's adaptive capabilities, we analyze the routing decisions of our final model (trained with $\lambda=0.15$) on each of the seven test benchmarks. Figure~\ref{fig:routing_distribution} visualizes the learned policy, revealing how the framework dynamically allocates resources based on the specific demands of each dataset.

The analysis highlights several key adaptive behaviors:
\begin{itemize}[itemsep=2pt, topsep=4pt]
    \item \textbf{Adaptation to Modality Strengths:} The model demonstrates a profound understanding of modality-task alignment. For \textbf{TABMWP}, a benchmark requiring structural and mathematical reasoning, the policy routes an overwhelming \textbf{97.2\%} of instances to the Image-only path. Conversely, for \textbf{InfoTabs}, a fact-verification task dependent on fine-grained semantics, the strategy shifts dramatically to favor the Text-only path (\textbf{67.5\%}).
    \item \textbf{Adaptation to Task Difficulty:} On challenging benchmarks where single-modality models are likely to fail, such as \textbf{TAT-QA}, the policy learns to prioritize correctness by invoking the powerful but costly Fusion path for \textbf{88.7\%} of cases. This demonstrates an ability to gauge task difficulty and escalate to a more robust strategy when necessary.
    \item \textbf{Balanced, Nuanced Strategies:} For benchmarks with a mix of challenges like \textbf{WTQ} and \textbf{FeTaQA}, the model learns a more balanced policy, distributing queries across all three paths. This indicates that the routing is not coarse-grained but makes fine-tuned, instance-level decisions.
\end{itemize}

In summary, this analysis provides compelling evidence that TableDART operates as a truly adaptive framework, which is the key to its state-of-the-art performance and efficiency.

\subsection{Effect of $\lambda$ on Inference Efficiency}
\label{sec:appendix_lambda_efficiency}

To more clearly characterize how the resource-aware objective influences efficiency, 
we report empirical latency and throughput measurements across different values of 
$\lambda$ in Table~\ref{tab:lambda_efficiency}. As shown, $\lambda$ has a direct 
impact on inference cost. While larger $\lambda$ generally encourages more efficient 
routing by discouraging unnecessary use of the Fusion path, the trend is not strictly 
monotonic. This is because each $\lambda$ induces a distinct routing policy that 
balances accuracy and computational cost differently.

\begin{figure}[ht!]
    \centering
    \includegraphics[width=\columnwidth]{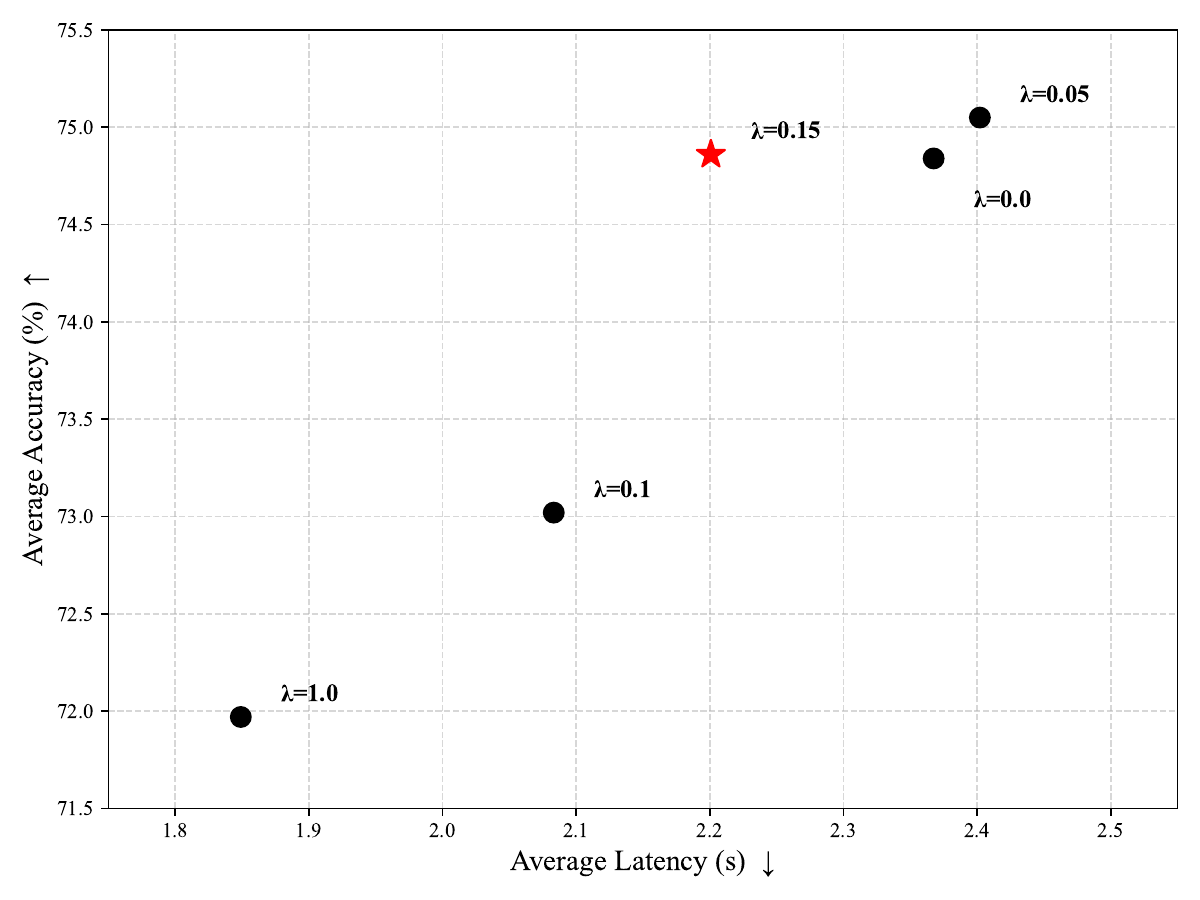} 
    \caption{Performance–efficiency trade-off across different values of the resource loss weight ($\lambda$). The red star marks the selected configuration.}
\label{fig:performance_efficiency_tradeoff}
\end{figure}

Figure~\ref{fig:performance_efficiency_tradeoff} visualizes this relationship by 
plotting average latency against average accuracy. The resulting curve illustrates 
a natural performance–efficiency frontier: smaller $\lambda$ values (e.g., $0.0$ 
and $0.05$) tend to overuse the Fusion path, while very large values (e.g., $1.0$) 
favor efficiency at the cost of reduced accuracy. The configuration $\lambda = 0.15$ 
provides the best overall balance, achieving the second-highest average accuracy (within 0.19 points of the best) while 
maintaining strong efficiency (8.4\% less inference latency).

% \clearpage

\clearpage
\begin{table}[h!]
\centering
\caption{Inference efficiency for different values of $\lambda$. 
Higher values encourage more efficient routing, though the effect is not strictly monotonic due to the distinct routing policies induced by each setting.}
\label{tab:lambda_efficiency}
\begin{tabular}{c S[table-format=1.4] S[table-format=2.2]}
\toprule
{$\lambda$} & {Latency (s)} & {Throughput (TPS)} \\
\midrule
0.00 & 2.3675 & 15.14 \\
0.05 & 2.4021 & 13.72 \\
0.10 & 2.0832 & 23.85 \\
0.15 & 2.2008 & 17.77 \\
1.00 & 1.8490 & 29.56 \\
\bottomrule
\end{tabular}
\end{table}

\end{document}